\documentclass[acmlarge, authorversion=True]{acmart}
\usepackage{url}
\usepackage{amsmath}
\usepackage{upgreek}
\usepackage{stmaryrd}
\usepackage{units}
\usepackage{bm}
\usepackage{relsize}
\usepackage{multirow}
\usepackage{graphics}
\usepackage{makecell}
\usepackage{enumitem}
\usepackage{amsfonts}
\usepackage{tabu}
\usepackage{balance}
\usepackage{setspace}
\usepackage{enumitem}

\AtBeginDocument{%
  \providecommand\BibTeX{{%
    \normalfont B\kern-0.5em{\scshape i\kern-0.25em b}\kern-0.8em\TeX}}}

\acmVolume{XX}
\acmNumber{XX}
\newtheorem{theorem}{Theorem}
\begin{document}

\title{Forecasting Graph-Based Time-Dependent Data with Graph Sequence Attention}

\author{Yang Li}
\orcid{0000-0002-9052-9308}
\affiliation{%
  \institution{Carnegie Mellon University}
  %\streetaddress{5000 Forbes Ave}
  \city{Pittsburgh}
  %\state{PA}
  \country{USA}
}
\affiliation{%
  \institution{Iowa State University}
  %\streetaddress{5000 Forbes Ave}
  \city{Ames}
  %\state{PA}
  \country{USA}
}
\email{jerryyangli@gmail.com}

\author{Di Wang}
\orcid{0000-0001-8003-9738}
\affiliation{%
  \institution{Microsoft}
  %\streetaddress{1 Microsoft Way}
  \city{Redmond}
  \country{USA}}
\email{diwangbruce@gmail.com}

\author{José~M.~F.~Moura}
\orcid{0000-0002-9822-8294}
\affiliation{%
  \institution{Carnegie Mellon University}
  %\streetaddress{5000 Forbes Ave}
  \city{Pittsburgh}
  %\state{PA}
  \country{USA}
}
\email{moura@ece.cmu.edu}

\renewcommand{\shortauthors}{Li, Wang, and Moura}

\begin{abstract}
Forecasting graph-based, time-dependent data has broad practical applications but presents challenges. Effective models must capture both spatial and temporal dependencies in the data, while also incorporating auxiliary information to enhance prediction accuracy. In this paper, we identify limitations in current state-of-the-art models regarding temporal dependency handling. 
To overcome this, we introduce \mbox{GSA-Forecaster}, a new deep learning model designed for forecasting in graph-based, time-dependent contexts. GSA-Forecaster utilizes graph sequence attention, a new attention mechanism proposed in this paper, to effectively manage temporal dependencies. \mbox{GSA-Forecaster} integrates the data's graph structure directly into its architecture, addressing spatial dependencies.
Additionally, it incorporates auxiliary information to refine its predictions further. We validate its performance using real-world graph-based, time-dependent datasets, where it demonstrates superior effectiveness compared to existing state-of-the-art models.
\end{abstract}

\begin{CCSXML}
<ccs2012>
<concept>
<concept_id>10010147.10010257.10010293</concept_id>
<concept_desc>Computing methodologies~Machine learning approaches</concept_desc>
<concept_significance>500</concept_significance>
</concept>
</ccs2012>
\end{CCSXML}

\ccsdesc[500]{Computing methodologies~Machine learning approaches}

\keywords{spatial dependency, temporal dependency, graph sequence attention, Transformer, GSA-Forecaster}

%\received{11 February 2024}
%\received[revised]{4 December 2024}
%\received[accepted]{22 February 2025}

\maketitle

\section{Introduction}
Many real-world applications deal with sets of dependent time series. These data exhibit two types of dependencies: intra-dependency, where data points within a single time series are correlated over time, and inter-dependency, where correlations exist across different time series. We term the former as \textit{temporal dependency} and the latter as \textit{spatial dependency}, a term we use regardless of the involvement of physical space. Such interconnected time series are pertinent in various fields such as economics~\cite{STS_Example_Economy}, climatology~\cite{STS_Example_Environment_Studies}, public health~\cite{STS_Example_Public_Health, TongShen_ICDE23}, transportation~\cite{DCRNN_ICLR2018}, cloud computing~\cite{zhang3, sizecap}, and Internet of things (IoT)~\cite{zhang1, zhang2}, among others. For instance, the hourly demand for taxi rides in different urban areas exemplifies these dependent time series, with demand at certain locations often peaking simultaneously. Similarly, sales volumes of diverse products in a retail setting represent dependent time series, where the sales of one product can affect those of others. In line with previous studies~\cite{Forecaster}, we model the spatial dependency of this data using a graph, referring to these interconnected time series as \textit{graph-based time-dependent data}.

\begin{comment}
\begin{figure}[ht]
\centering
\includegraphics[width=0.55\columnwidth]{figures/graph_data_definition.pdf}
\caption{Definition of graph-based time-dependent data.}
\label{fig:definition_of_data}
\end{figure}
\end{comment}

\begin{figure}[t]
\centering
\begin{minipage}[b]{0.5\columnwidth}
    \centering
    \includegraphics[width=\textwidth]{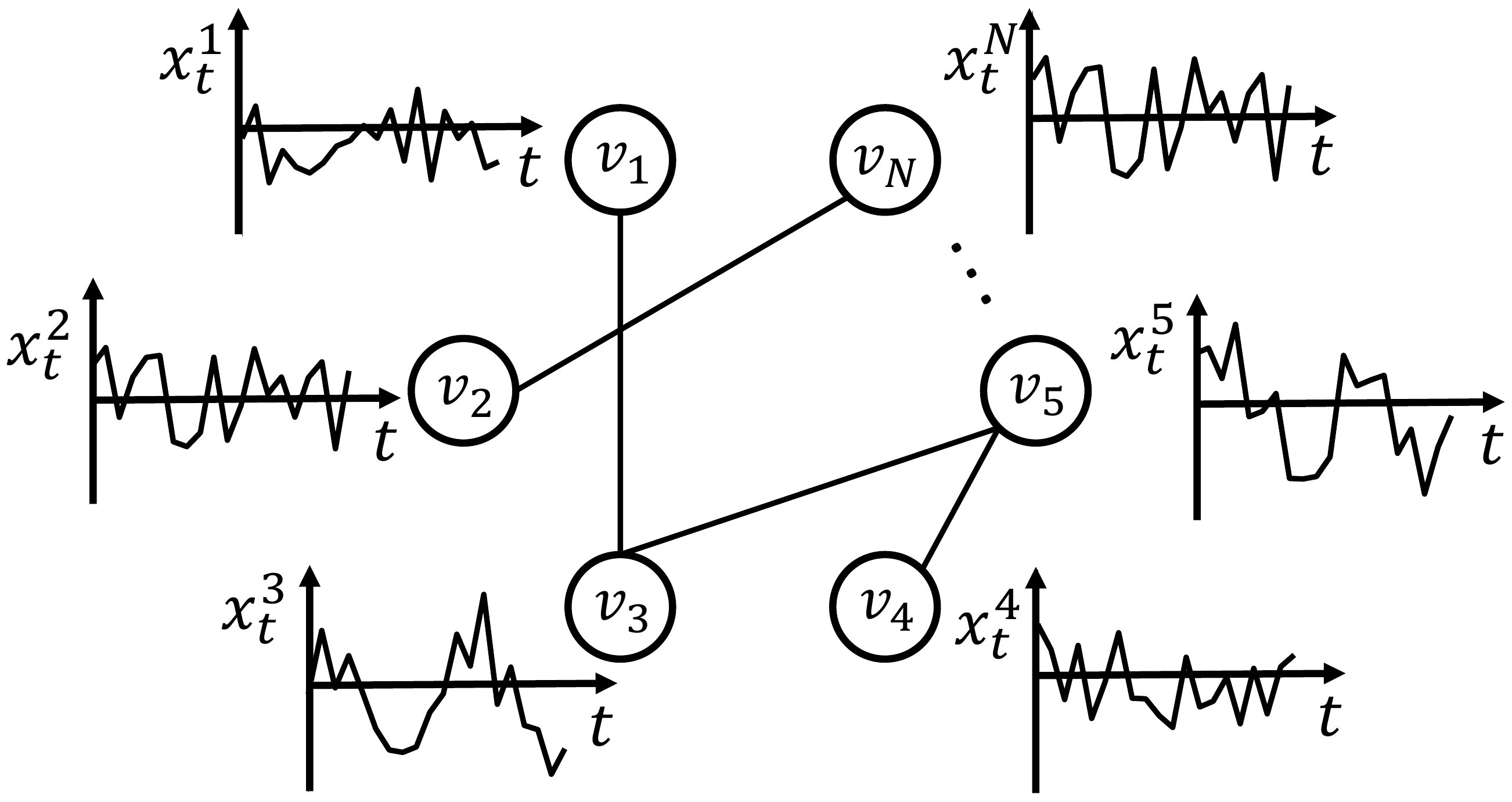}
    \captionof{figure}{Definition of graph-based time-dependent data.}
    \label{fig:definition_of_data}
\end{minipage}
\hfill
\begin{minipage}[b]{0.44\columnwidth}
    \centering
    \includegraphics[width=\textwidth]{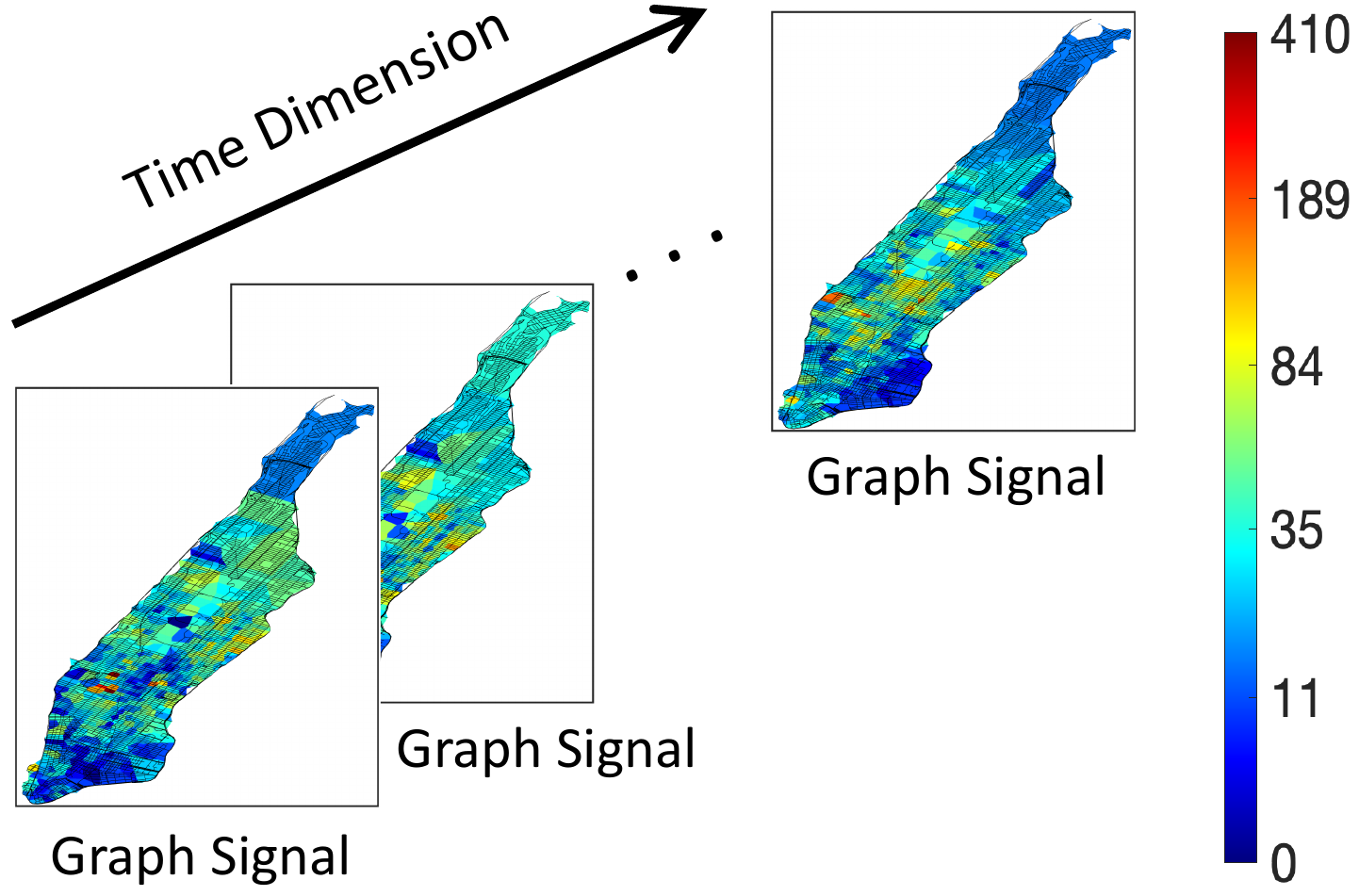}
    \captionof{figure}{Example of graph-based time-dependent data.}
    \label{fig:example_of_data}
\end{minipage}
\end{figure}

Figure~\ref{fig:definition_of_data} illustrates how we formalize this type of data. In its underlying graph $G$, each node represents a time series, while an edge indicates a spatial dependency between two series.
We define \textit{graph-based time-dependent data} as a sequence of graph signals $\left\{ \mathbf{x}_{t}\mid\mathbf{x}_{t}\in\mathbb{R}^{N},\:t=1,\ldots,T\right\}$. Each \textit{graph signal} $\mathbf{x}_{t}$ encapsulates the data $x_{t}^{i}$ at every node $v_{i}$  in graph $G$ for a specific time $t$, meaning $\mathbf{x}_{t}=\left[{x_{t}^{1}}, \cdots, {x_{t}^{N}}\right]^{\top}$. Here, $N$ denotes both the number of time series and the number of nodes in graph $G$.
One way to visualize graph signal $\mathbf{x}_{t}$ is through a heatmap. For instance, a heatmap displaying hourly taxi demand across various locations at a specific time represents a graph signal. In this scenario, the graph illustrates the spatial dependency of taxi demands at different city locations. A collection of such heatmaps, as depicted in Figure~\ref{fig:example_of_data}, constitutes graph-based time-dependent data.\footnote{Graph-based time-dependent data is a type of multivariate time series that features strong and stable dependencies among some, but not all, of its dimensions. A specific category within graph-based time-dependent data is known as \textit{spatial time series}~\cite{Spatial_Time_Series_Def_2003,Spatial_Time_Series_Def_2019}, or \textit{spatial- and time-dependent data}~\cite{Forecaster}, where the graph depicts relationships between data points at different physical locations. However, not all graph-based time-dependent data is linked to physical locations; for example, consider the sales volumes of various products. Graph-based time-dependent data differs from \textit{temporal graphs}~\cite{temporal_graph}, another graph data variant. In temporal graphs, nodes do not carry associated data; instead, only the edges of the graph evolve over time.}

\begin{comment}
\begin{figure}
\centering
\includegraphics[width=0.48\columnwidth]{figures/graph_data_example.pdf}
\caption{Example of graph-based time-dependent data: a series of hourly taxi ride-hailing demand heatmaps.}
\label{fig:example_of_data}
\end{figure}
\end{comment}

Forecasting graph-based time-dependent data is crucial for numerous practical applications that significantly impact our lives and society. However, this task is challenging as it requires models to accurately capture both \textit{spatial and temporal dependencies} in the data while also considering relevant \textit{auxiliary information} for precise predictions. Firstly, in many cases, the graph structure representing spatial dependency among different time series is unknown. Effective models should not only account for this structure but also exploit it to enhance forecasting accuracy. Secondly, graph-based time-dependent data can exhibit both \textit{short-range} and \textit{long-range} temporal dependencies. For instance, data for the next time step might be influenced not just by recent data (short-range dependency) but also by data from much earlier periods (long-range dependency). This complexity is further amplified by \textit{non-stationarity}, where unforeseen events alter the temporal relationships between upcoming and historical data. Models need to discern and focus on the specific historical data that is most relevant to future predictions. Thirdly, auxiliary information, such as weather conditions affecting taxi demand, can significantly influence the evolution of this data. Incorporating such auxiliary data is vital for more accurate forecasting.

Traditional forecasting models like auto-regressive integrated moving average (ARIMA)~\cite{VAR} and more recent methods like vector autoregression (VAR)~\cite{VAR} and causal graph processes~\cite{CausalGraphProcess} often assume data stationarity, an assumption that is frequently not valid~\cite{DCRNN_ICLR2018, Spatio_Temporal_ICDM_2017}. To tackle the non-stationarity and complex nature of the data, deep learning-based models have been introduced~\cite{DCRNN_ICLR2018,DMVSTNET_AAAI2018,jain2016structural,STGCN_IJCAI2018,STMGCN_AAAI2019,STResNet_AAAI17,Yao_AAAI2019,Spatio_Temporal_ICDM_2017, zhang1, zhang2, zhang3, GraphWaveNet, AGCRN}. These models typically employ recurrent neural networks (RNN), convolutional neural networks (CNN), deep belief networks (DBN), or their variants to grasp temporal dependencies. However, they may fall short in capturing long-range temporal dependencies across distant time points~\cite{long_range_dependency_rnn, Transformer}. To address this, the Forecaster model~\cite{Forecaster}, based on an attention mechanism, has been developed. Built upon the Transformer architecture, which utilizes attention to capture long-range dependencies, Forecaster extends this concept for graph-based time-dependent data. It incorporates the graph structure into its design to account for spatial dependencies and uses the theory of Gaussian Markov random fields~\cite{Gaussian_Markov_Book} to deduce unknown graph structures. Forecaster has shown state-of-the-art results in predicting graph-based time-dependent data~\cite{Forecaster}, proving its effectiveness in handling these complex datasets.

\begin{figure}[t]
\centering
\begin{minipage}[b]{0.48\columnwidth}
    \centering
    \includegraphics[width=\textwidth]{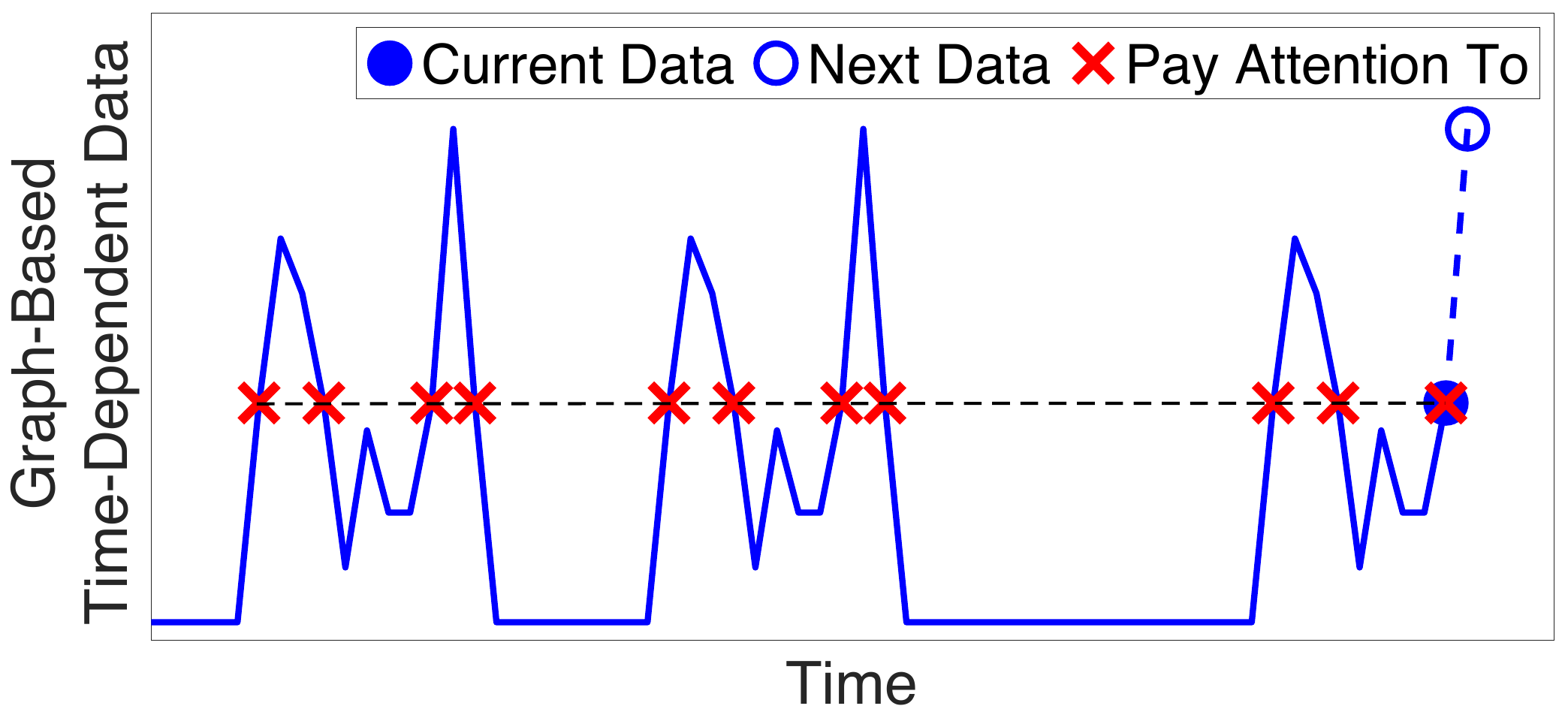}
    \captionof{figure}{Illustration of limitations of standard attention. 
    In this example, there is only a single node in the graph structure.}
    \label{fig:standard_intro}
\end{minipage}
\hfill
\begin{minipage}[b]{0.46\columnwidth}
    \centering
    \includegraphics[width=\textwidth]{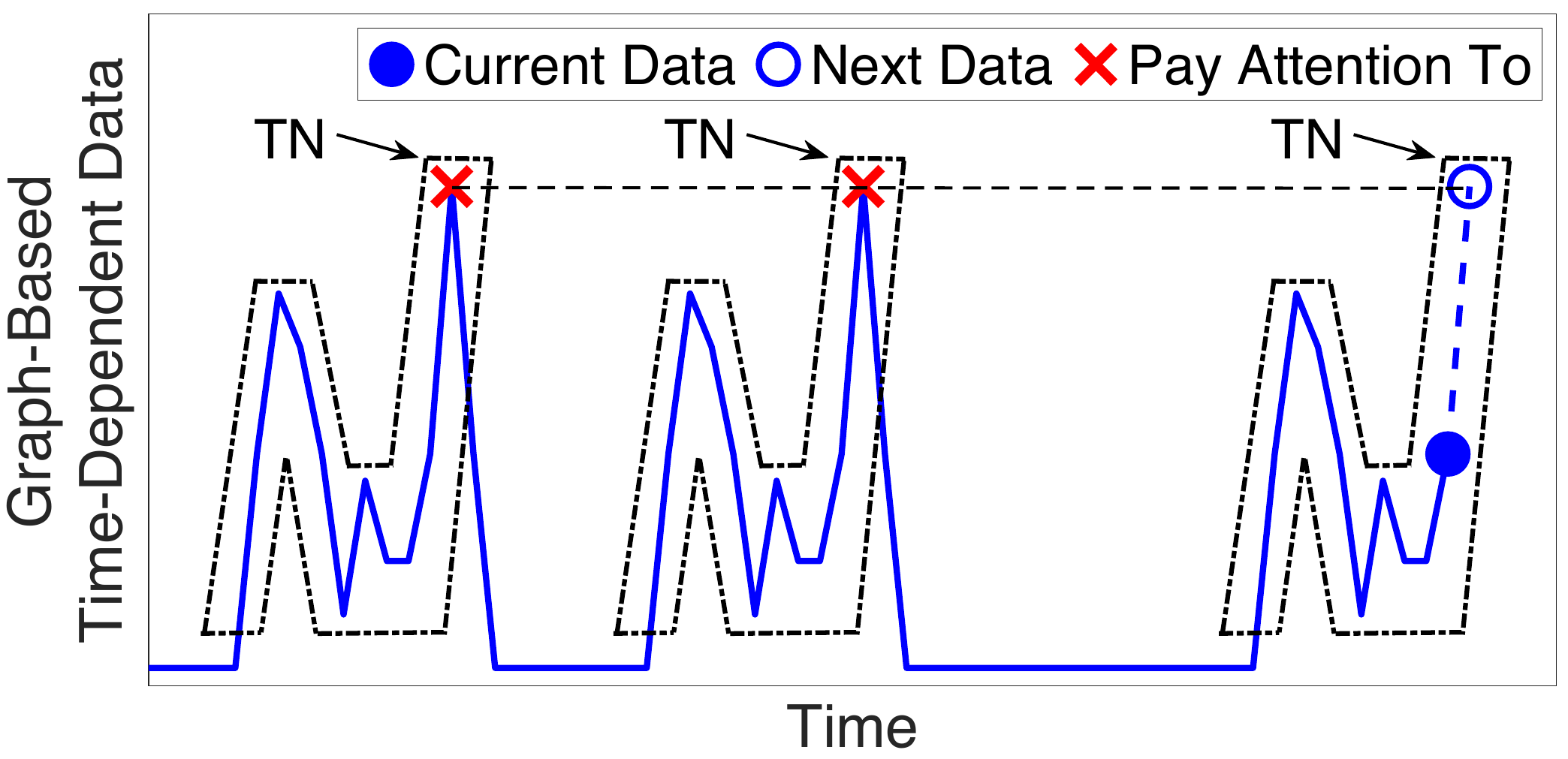}
    \captionof{figure}{Illustration of Graph Sequence Attention and 
    how it addresses the limitation of standard attention.}
    \label{fig:gsa_intro}
\end{minipage}
\end{figure}

Although Forecaster has achieved impressive results, we note that its standard attention mechanism is not entirely adequate for capturing temporal dependencies in graph-based time-dependent data. For instance, as depicted in Figure~\ref{fig:standard_intro}, to forecast the next data point (represented by a blue empty circle), Forecaster initially uses the current data point (illustrated by a blue dot) as its preliminary estimate.\footnote{In this context, \textit{data point} at a time instant is synonymous with \textit{graph signal} at that time instant. Throughout this paper, we use these terms interchangeably.} Following this, Forecaster employs the standard attention mechanism to compare this initial estimate with each piece of historical data, focusing on those that are similar (marked by red crosses). Consequently, Forecaster relies on historical data points resembling the current one for its predictions. However, this approach can lead to significant inaccuracies due to two main reasons: 1) a substantial disparity between the current and the next data points often renders the initial estimate, based on current data, highly unreliable; and 2) Forecaster tends to focus on inappropriate historical data points (such as the red crosses) because it identifies them based solely on their similarity to the current data, leading to a misinterpretation of the actual temporal dependency.\footnote{The standard attention mechanism is notably effective in NLP tasks, where consecutive words often bear a close semantic relationship. However, this is not as straightforward for graph-based time-dependent data consisting of numerical values. In such datasets, data points at successive time instants can vary greatly, particularly in the presence of non-stationarity. This significant variation renders the standard attention mechanism less effective in identifying temporal dependencies in data characterized by numerical values.}

To overcome the limitations of standard attention mechanisms in capturing temporal dependency, we introduce \textit{graph sequence attention} (GSA).\footnote{We name our proposed attention mechanism \textit{graph sequence attention} because it conducts attention-related operations specifically on graph signals.} GSA addresses the issue inherent in standard attention --- relying on a potentially inaccurate initial estimate to compare with each piece of historical data, which can lead to misguided focus. GSA overcomes this by utilizing a cluster of more recent data points, providing a more accurate representation of temporal dependency. We explain the workings of GSA using Figure~\ref{fig:gsa_intro}. In this illustration, GSA aggregates several recent data points with the initial estimate, creating a \textit{temporal neighborhood} (TN) for the forthcoming time instant. Similarly, for each piece of historical data, GSA forms a \textit{historical temporal neighborhood} by grouping it with an equivalent number of preceding data points. GSA then compares the TN for the upcoming time instant against each historical temporal neighborhood. This process involves multiple comparisons of their respective data points --- comparing the most recent, the second most recent, and so on, down to the earliest. Upon identifying similar historical temporal neighborhoods, GSA focuses on their most recent data points and utilizes these values for predicting the upcoming data point. By employing temporal neighborhoods for comparison --- rather than individual data points, as standard attention does --- GSA is more adept at managing individual data point inaccuracies and noise (e.g., errors in the initial estimate), leading to more effective and accurate attention in forecasting.

We introduce \textit{GSA-Forecaster}, a new deep learning model for forecasting graph-based time-dependent data, leveraging our graph sequence attention mechanism. This model uses graph sequence attention to effectively capture temporal dependencies. Like its predecessor, Forecaster~\cite{Forecaster}, \mbox{GSA-Forecaster} integrates the graph structure into its architecture to encode spatial dependencies and also accounts for essential auxiliary information. We applied GSA-Forecaster to the task of predicting taxi ride-hailing demand in Manhattan, New York City. The city was divided into 471 regions, and our goal was to forecast the hourly taxi demand in these regions. A graph representing the relationships between these locations was constructed using a method suggested by Forecaster. Our evaluation was conducted using the NYC Taxi dataset~\cite{NYC_Taxi}, which includes all taxi trips in Manhattan from January 1, 2009, to June 30, 2016 — a total of over 1.06 billion trips. The results show that GSA-Forecaster outperforms existing predictors, reducing the root-mean-square error (RMSE) by 6.7\% and the mean absolute percentage error (MAPE) by 5.8\% compared to Forecaster. Additionally, GSA-Forecaster was applied to predict traffic speeds measured at each of 325 sensors in the San Francisco Bay Area, as per the \mbox{PEMS-BAY} dataset~\cite{DCRNN_ICLR2018} covering the period from January 1, 2017, to June 30, 2017. In this case, GSA-Forecaster achieved higher forecasting accuracy than other state-of-the-art predictors. We further assess the performance of GSA-Forecaster by applying it to predict electricity consumption loads (using the ECL dataset~\cite{informer_ecl}) and road occupancy rates (using the Traffic dataset~\cite{autoformer_traffic}). Our results demonstrate that GSA-Forecaster performs better in comparison to state-of-the-art predictors.

%The rest of the paper is organized as follows. Section \ref{sec:challenges} uses real-world data to illustrate the challenges of forecasting graph-based time-dependent data. Section \ref{sec:forecaster} gives a brief overview of Forecaster and analyzes the limitation of the standard attention mechanism. Section \ref{sec:gsa} introduces our proposed graph sequence attention and GSA-Forecaster. We review related work in Section \ref{sec:related} and draw conclusions in Section \ref{sec:conclusion}. 

This paper makes the following major contributions:
\begin{itemize}[topsep=2pt]
\item \textbf{We analyze real-world datasets}: We illustrate the crucial need to capture both spatial and temporal dependencies and to consider auxiliary information when forecasting graph-based, time-dependent data.
    \item \textbf{We propose Graph Sequence Attention:} A new attention mechanism that leverages temporal neighborhood information for multiple comparisons. This effectively identifies historical data that better captures temporal dependency, overcoming the limitations of the standard attention mechanism found in models such as Transformer and Forecaster.
    \item \textbf{We develop GSA-Forecaster:} A new deep learning model designed for forecasting graph-based time-dependent data. This model captures spatial and temporal dependencies and accounts for auxiliary information.
    \item \textbf{We apply GSA-Forecaster to real-world datasets:} The model is tested on both the large-scale NYC Taxi dataset and the medium-scale PEMS-BAY, ECL, and Traffic datasets. Through these applications, we demonstrate GSA-Forecaster's effectiveness and superiority over previous methods.
\end{itemize}

\section{\label{sec:challenges}Problem Statement and Challenges in Forecasting Graph-based Time-Dependent Data}
In this section, we formalize the problem of forecasting graph-based, time-dependent data and highlight the associated challenges, using the hourly taxi ride-hailing demand in New York City~\cite{NYC_Taxi} as an illustrative example. Overcoming these challenges is essential for achieving accurate forecasting models.

\subsection{\label{sec:task_gsa}Forecasting Task}
The forecasting task involves predicting $T'$ future graph signals based on $T$ historical graph signals, alongside $T+T'$ historical and future auxiliary information. The task is framed as learning a function, $f(\cdot)$, that performs the following mapping:

\begin{equation}
\left\{
\begin{array}{c}
\left\{\mathbf{x}_{t-T+1},\cdots,\mathbf{x}_{t}\right\} \\
\left\{\mathbf{a}_{t-T+1},\cdots,\mathbf{a}_{t+T'}\right\} 
\end{array}
\right\} \overset{f\left(\cdot\right)}{\longrightarrow}
\left\{\mathbf{x}_{t+1},\cdots,\mathbf{x}_{t+T'}\right\} 
\label{eq:task}
\end{equation}
Here, $\mathbf{x}_{t}$ represents the graph signal at time $t$, defined as $\mathbf{x}_{t} = \left[{x_{t}^{1}}, \cdots, {x_{t}^{N}}\right]^{\top} \in \mathbb{R}^{N}$, where $x_{t}^{i}$ denotes the data at node $i$ at time $t$. The auxiliary information at time $t$ is denoted by $\mathbf{a}_{t}$, belonging to the space $\mathbb{R}^{A}$, where $A$ represents the dimensionality of the auxiliary data.\footnote{For simplicity, we assume that different nodes share the same auxiliary information. Generalizing our model to other cases is straightforward.}

\subsection{Challenges: Spatial Dependency}
\begin{figure}[h]
\centering
\includegraphics[width=0.55\columnwidth]{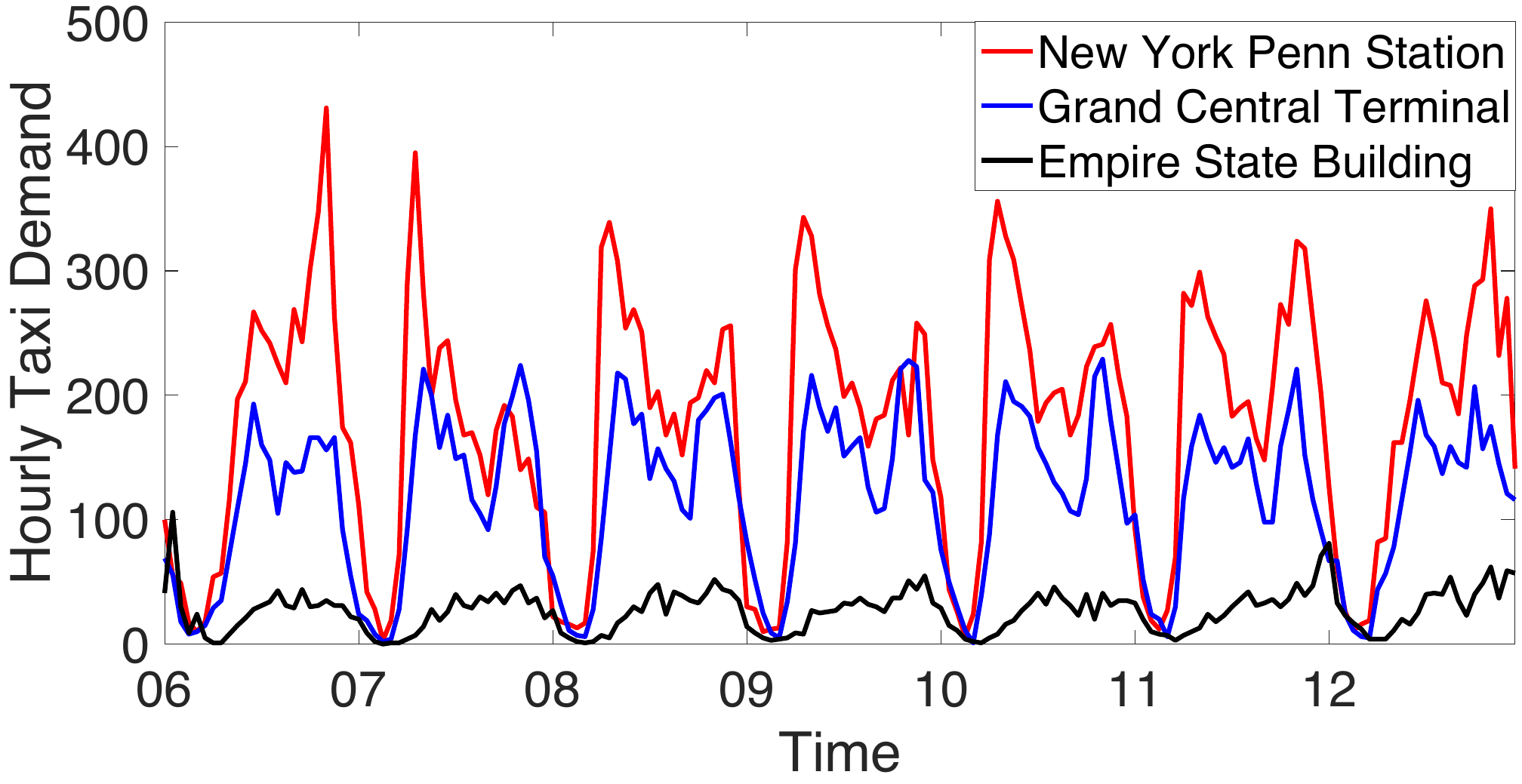}
\caption{llustrating spatial dependency --- This chart presents the hourly taxi demand near New York Penn Station, Grand Central Terminal, and the Empire State Building, covering the period from 12:00 a.m. on Sunday, March 6, 2016, to 11:59 p.m. on Saturday, March 12, 2016. The x-axis tick labels indicate the start of each date (for instance, `06' denotes the beginning of March 6, 2016, at 12:00 a.m.).}
\label{fig:spatial_dependency}
\end{figure}

The first challenge is modeling spatial dependency, which is inherent in graph-based time-dependent data comprising interrelated time series. This is exemplified by the hourly taxi demand near New York Penn Station and Grand Central Terminal in Manhattan. Figure~\ref{fig:spatial_dependency} shows that changes in taxi demand at one location often reflect similar changes at the other, evidenced by a 0.82 Pearson correlation coefficient. Spatial dependency can be short-range, occurring between adjacent locations, or long-range, between distant locations, and is not strictly related to physical distance. For example, despite the larger physical distance between New York Penn Station and Grand Central Terminal compared to their distance from the Empire State Building, their taxi demands are more strongly correlated (indicating long-range spatial dependency) than their correlations with the Empire State Building (0.33 and 0.48, showing weaker spatial dependencies). Leveraging both short-range and long-range spatial dependencies is essential for precise forecasting.

\subsection{Challenges: Temporal Dependency}
\begin{figure}[h]
\centering
\includegraphics[width=0.58\columnwidth]{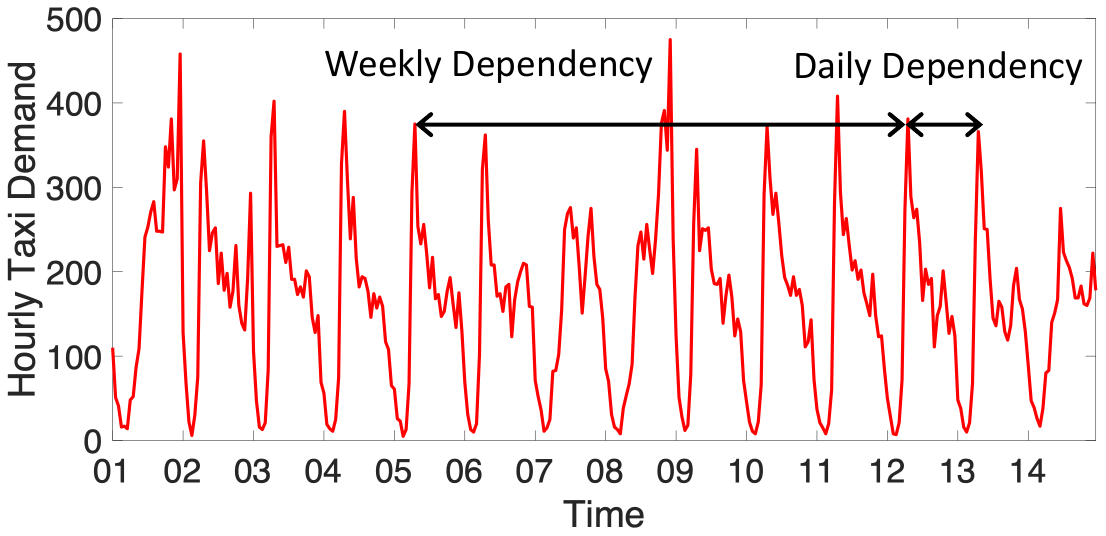}
\caption{Illustrating temporal dependency --- This chart depicts the hourly taxi demand near New York Penn Station, from 12:00 a.m. on Sunday, May 1, 2016, to 11:59 p.m. on Saturday, May 14, 2016. The x-axis tick labels indicate the start of each date (for example, `01' represents the beginning of May 1, 2016, at 12:00 a.m.).}
\label{fig:temporal_dependency}
\end{figure}

The second challenge concerns capturing temporal dependency, which refers to the relationships between data at different time points. This dependency can manifest as short-range (e.g., daily patterns) or long-range (e.g., weekly patterns). For instance, hourly taxi demand near New York Penn Station demonstrates both daily and weekly temporal dependencies (Figure~\ref{fig:temporal_dependency}). Accurately capturing both types of temporal dependency is vital for effective forecasting.

\subsection{Challenges: Auxiliary Information}
\begin{figure}[h]
\centering
\includegraphics[width=0.52\columnwidth]{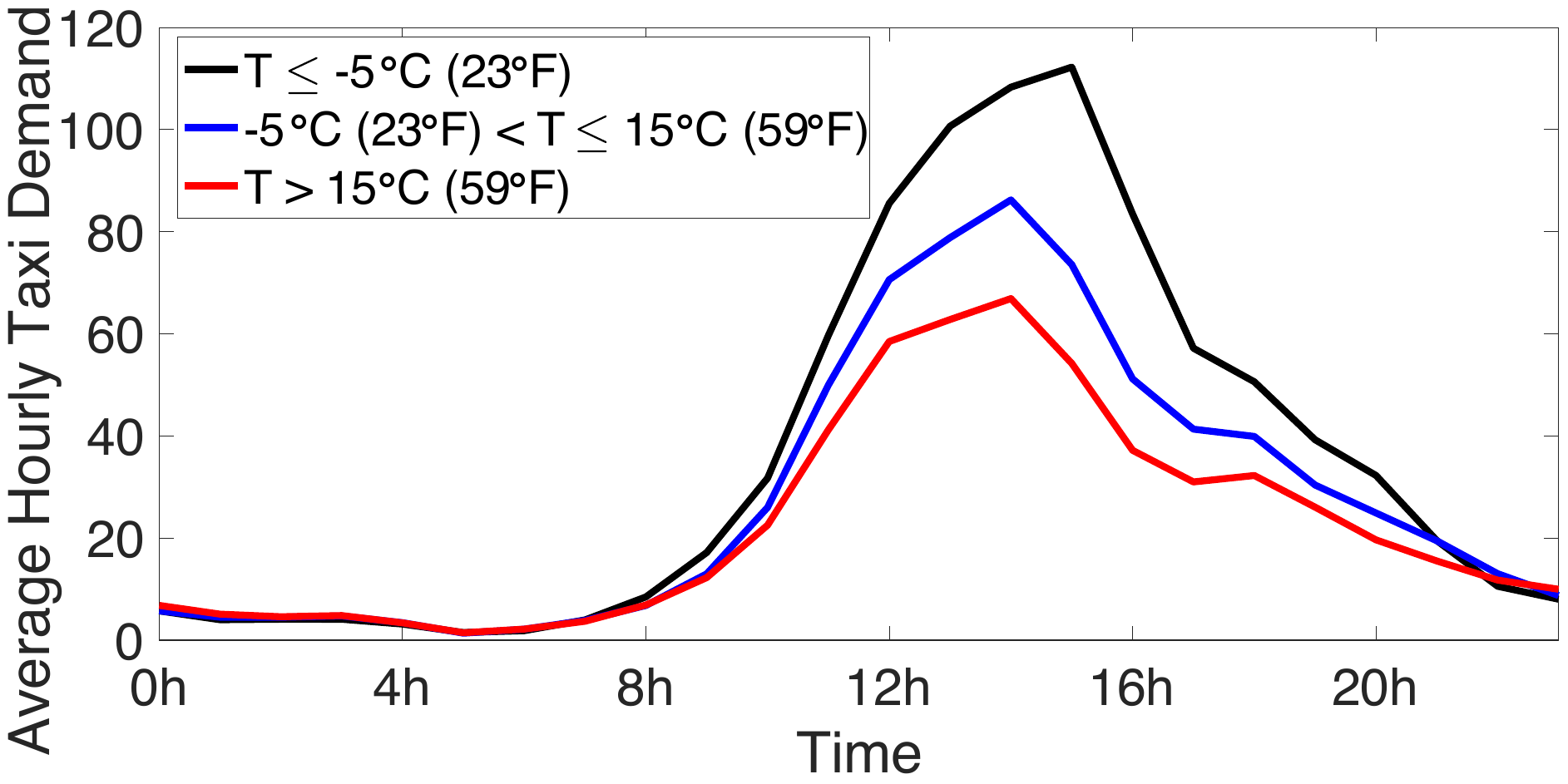}
\caption{Illustrating auxiliary information --- This graph shows the effect of temperature ($T$) on the average hourly taxi demand near the Javits Center, across various hours ($h$) of the day, from January 1, 2009, to June 30, 2016.}
\label{fig:temperature_impact}
\end{figure}

The third challenge involves incorporating auxiliary information that can enhance the data used for forecasting. An example is the impact of weather on taxi demand around the Javits Center, a major convention center in New York City. Figure~\ref{fig:temperature_impact} shows how temperature variations influenced the average hourly taxi demand across various hours of the day. Notably, taxi demand increases when temperatures drop, possibly because colder weather encourages people to opt for taxis. Incorporating such significant external factors like weather, which may not be directly reflected in the primary data, is crucial for enhancing the accuracy of forecasting.

\section{\label{sec:gsa}GSA-Forecaster \& Graph Sequence Attention}
In this section, we present GSA-Forecaster and graph sequence attention. We begin by introducing how to identify the graph structure of the data (Section~\ref{sec:graph}). Next, we present the architecture of GSA-Forecaster (Section~\ref{sec:arch_gsa}). We then detail the central element of our architecture, the proposed graph sequence attention (Sections~\ref{sec:gsa_decoding} and \ref{sec:gsa_encoding}). Finally, we introduce node-level attention as a solution for handling highly non-stationary scenarios (Section~\ref{sec:node_level_attention}).

\subsection{\label{sec:graph} How to Identify Graph}
For graph-based, time-dependent data, if a graph is provided, such as in the PEMS-Bay dataset, we can directly leverage this graph. If no graph is provided, such as in the NYC Taxi, ECL, and Traffic datasets, we can use Gaussian Markov random fields~\cite{Gaussian_Markov_Book} to identify the graph structure.

Gaussian Markov random fields model a graph signal $\mathbf{x}_t$, which consists of the data $x_{t}^{i}$ at each node $v_{i}$ at a given time $t$, i.e., $\mathbf{x}_{t}=\left[{x_{t}^{1}}, \cdots, {x_{t}^{N}}\right]^{\top}$, as a random vector drawn from a multivariate Gaussian distribution. The probability density function of $\mathbf{x}_t$ is given by:
\begin{equation}
f\left(\mathbf{x}_t\right)=\frac{\left|Q\right|}{\left(2\pi\right)^{\nicefrac{N}{2}}}\exp\left(-\frac{1}{2}\left(\mathbf{x}_t-\bm{\mu}\right)^{\top}Q\left(\mathbf{x}_t-\bm{\mu}\right)\right)
\end{equation}
where $\bm{\mu}$ and $Q$ are the mean and precision matrix of the distribution.

The precision matrix \( Q \) characterizes the relationships between data at different nodes and can be estimated using the graphical lasso~\cite{Graphical_Lasso}, a sparse-penalized maximum likelihood estimator. Once \( Q \) is estimated, GSA-Forecaster leverages it to compute the conditional correlation between the data at every pair of nodes~\cite{gu2007learning, Forecaster}:
\begin{equation}
\text{Corr} \left( x_{t}^{i}, x_{t}^{j} \mid x_{t}^{-ij} \right) = -\frac{Q_{ij}}{\sqrt{Q_{ii}Q_{jj}}}
\end{equation}
where \(\text{Corr} \left( x_{t}^{i}, x_{t}^{j} \mid x_{t}^{-ij} \right)\) represents the conditional correlation between the data at nodes \(i\) and \(j\), and \(Q_{ij}\) denotes the entry at the \(i^\textrm{th}\) row and \(j^\textrm{th}\) column of the precision matrix \(Q\). A large absolute value of the conditional correlation implies a spatial dependency between the two nodes, thereby introducing an edge to connect them.

\subsection{\label{sec:arch_gsa}Architecture of GSA-Forecaster}
GSA-Forecaster operates in an autoregressive manner for multi-step forecasting. After predicting the future graph signals $\{\mathbf{x}_{t+i}\mid i=1,\ldots,k-1\}$, it utilizes these predictions, in conjunction with other known information, to forecast the next graph signal $\mathbf{x}_{t+k}$. This process is repeated iteratively for $k=1,\ldots,T'$, allowing the model to predict all $T'$ future graph signals. Through this method, GSA-Forecaster effectively transforms the multi-step forecasting task into a sequence of single-step forecasting tasks for $\mathbf{x}_{t+k}$.

In the rest of Section \ref{sec:gsa}, we delve deeper into how GSA-Forecaster tackles these single-step forecasting challenges. For consistency, we use $t$ to denote the current timestamp, $\widehat{\mathbf{x}}_{t+k}$ for our forecasted results for $\mathbf{x}_{t+k}$, and $\{\mathbf{x}_{t+i}\mid i=1,\;\ldots,\;k-1\}$ to represent the previously predicted graph signals.\footnote{In the remainder of this section, we do not engage with the \textit{actual values} of the graph signals $\{\mathbf{x}_{t+i}\mid i=1,\;\ldots,\;k-1\}$. To simplify our equations, we employ the same notations for both the \textit{actual} and \textit{predicted} values of these graph signals.}

\begin{figure*}[t]
\centering
\includegraphics[width=1.0\columnwidth]{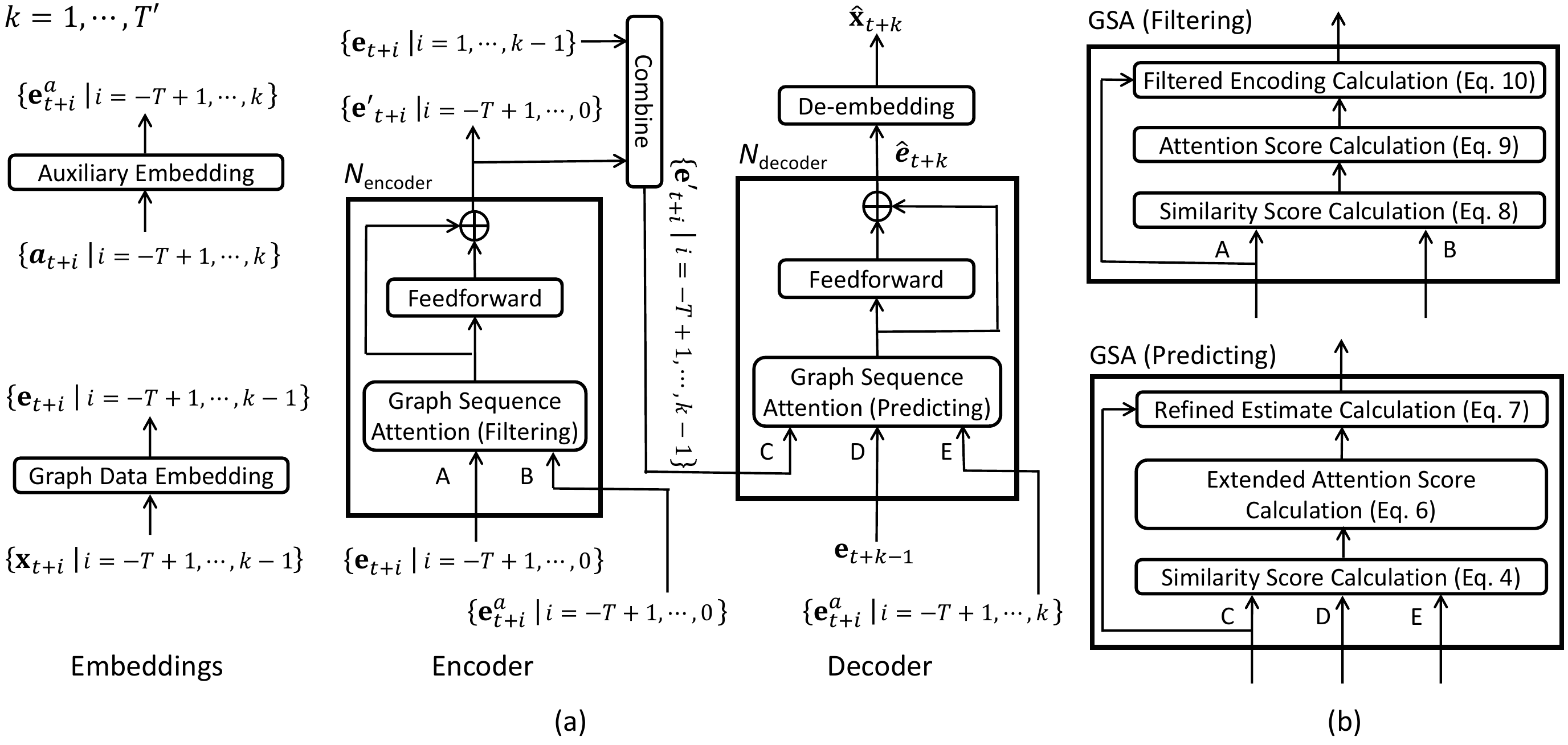}
\caption{Diagram of (a) GSA-Forecaster and (b) graph sequence attention (GSA).} 
\label{fig:GSA_Forecaster}
\end{figure*}

The architecture of GSA-Forecaster, as illustrated in Figure~\ref{fig:GSA_Forecaster}(a), comprises \textit{embeddings}, an \textit{encoder}, and a \textit{decoder}. Both the encoder and the decoder utilize two distinct types of graph sequence attention --- \textit{GSA (Filtering)} for noise reduction and \textit{GSA (Predicting)} for data forecasting --- as depicted in Figure~\ref{fig:GSA_Forecaster}(b). In line with Forecaster~\cite{Forecaster}, \mbox{GSA-Forecaster} utilizes sparse linear layers, a variant of graph neural networks, to model spatial dependencies. Once the graph structure is established, all linear layers in the model are explicitly designed to reflect it. Specifically, each linear layer assigns neurons to graph nodes to encode their respective data. Neurons corresponding to connected nodes (indicating a dependency) are also connected to capture this relationship, while neurons for unconnected nodes remain independent. Thus, all linear layers in the model function as sparse layers, extending this structural adaptation beyond embedding layers to also include attention layers. Furthermore, \mbox{GSA-Forecaster} is highly adaptable and can be integrated with various types of graph neural networks.

\subsubsection{Embeddings}
GSA-Forecaster features two key embedding components: graph data embedding and auxiliary embedding. These components produce embedding representations for graph data and auxiliary information, respectively, aiding the encoder and decoder in forecasting future graph signals. The graph data embedding employs sparse linear layers and nonlinear activation functions to encode each graph signal $\mathbf{x}_{t+i}$ into an encoding $\mathbf{e}_{t+i}\in\mathbb{R}^{d_{\mathrm{model}}}$, where $i$ ranges from 
$-T+1$ to $k-1$. Similarly, the auxiliary embedding encodes each piece of
auxiliary information $\mathbf{a}_{t+i}$ into
$\mathbf{e}_{t+i}^{a}\in\mathbb{R}^{d_{\mathrm{aux}}}$, over the same range of $i$.
\subsubsection{Encoder}
The encoder is designed to refine encodings of historical graph signals. Given that random events can introduce noise into these signals, the encoder's role is to mitigate this noise, enabling more accurate predictions based on cleaner historical data. It processes the encodings of historical graph signals 
$\left\{\mathbf{e}_{t+i}\mid i=-T+1,\ldots,0\right\}$ and corresponding auxiliary information $\left\{\mathbf{e}_{t+i}^{a}\mid i=-T+1,\ldots,0\right\}$
through $N_{\mathrm{encoder}}$ encoder layers. Each layer includes a 
\mbox{\textit{GSA (Filtering)}} layer and a feedforward neural network, crucial for noise filtering and capturing spatial dependencies, respectively.

\subsubsection{Decoder}
Tasked with predicting the next graph signal, the decoder commences with the encoding of the last predicted graph signal, $\mathbf{e}_{t+k-1}$, as its initial estimate. It combines this estimate with a series of previous data encodings --- both filtered historical graph signals $\{\mathbf{e}_{t+i}'\mid i=-T+1,\ldots,0\}$ and encodings of prior predicted signals $\{\mathbf{e}_{t+i}\mid i=1,\ldots,k-1\}$. Taking these inputs, along with relevant auxiliary information $\{\mathbf{e}_{t+i}^a\mid i=-T+1,\ldots,k\}$, the decoder passes them through $N_{\mathrm{decoder}}$ decoder layers to produce the final estimated encoding of the next graph signal, $\widehat{\mathbf{e}}_{t+k}$. Each layer consists of a \textit{graph sequence attention (predicting)} layer and a feedforward network, which together refine the prediction by considering both temporal and spatial dependencies.

Finally, a de-embedding layer decodes this final estimate, $\widehat{\mathbf{e}}_{t+k}$, to predict the next graph signal, 
$\widehat{\mathbf{x}}_{t+k}$. This layer uses sparse linear layers and nonlinear activation functions to effectively translate the encoding back into a graph signal format.

\subsection{\label{sec:gsa_decoding}Graph Sequence Attention (Predicting)}
The primary function of GSA (predicting) within each decoder layer is to forecast the encoding for the upcoming graph signal. It achieves this by identifying temporally related previous graph signals to aid in predicting the next one. Inputs for this process, as illustrated in Figure~\ref{fig:GSA_Forecaster}(b), include:
\begin{itemize}
    \item Entry C: Previous data sequence $\left\{\mathbf{e}_{t+i}'\mid i=-T+1,\ldots,k-1\right\}$
    \item Entry D: Estimated encoding of the next graph signal\footnote{For the first decoder layer, the `Entry D' input, denoted as $\mathbf{e}_{t+k}'$, is set to the initial estimate $\mathbf{e}_{t+k-1}$. In contrast, for subsequent decoder layers, this `Entry D' input $\mathbf{e}_{t+k}'$ is updated to reflect the refined estimate produced by the preceding decoder layer. This refined estimate may differ from the initial estimate, allowing for progressive enhancement of the prediction accuracy as it passes through successive layers.}  $\mathbf{e}_{t+k}'$

    \item Entry E: Auxiliary information $\left\{\mathbf{e}_{t+i}^a\mid i=-T+1,\ldots,k\right\}$
\end{itemize}
The output is a \textit{refined} estimate for the encoding of the next graph signal, denoted as $\widehat{\mathbf{e}}_{t+k}$. The process to derive $\widehat{\mathbf{e}}_{t+k}$ involves several steps:

Initially, GSA (predicting) calculates the \textit{similarity scores} between the forthcoming graph signal $\mathbf{x}_{t+k}$ and each previous signal  $\mathbf{x}_{t+i}$. With these similarity scores, we can find the previous graph signals that the next graph signal truly depends on. These scores are determined using the formula:
\begin{equation}
\begin{array}{lll}
s_{t+k,\,t+i}^{(h)}  =   \dfrac{1}{M}{\mathlarger{\sum}\limits_{m=0}^{M-1}w\left\langle {\frac{W_{Q}^{(h)}\mathbf{e}'_{t+k-m}}{\left|W_{Q}^{(h)}\mathbf{e}'_{t+k-m}\right|}},\; {\frac{W_{K}^{(h)}\mathbf{e}'_{t+i-m}}{\left|W_{K}^{(h)}\mathbf{e}'_{t+i-m}\right|}}\right\rangle}
    +\, w_{A}\left\langle \frac{W_{QA}^{(h)}\mathbf{e}_{t+k}^{a}}{\left|W_{QA}^{(h)}\mathbf{e}_{t+k}^{a}\right|},\;\frac{W_{KA}^{(h)}\mathbf{e}_{t+i}^{a}}{\left|W_{KA}^{(h)}\mathbf{e}_{t+i}^{a}\right|}\right\rangle 
    +\, w_{P}\left\langle \frac{W_{QP}^{(h)}\mathbf{e}_{k}^{p}}{\left|W_{QP}^{(h)}\mathbf{e}_{k}^{p}\right|},\;\frac{W_{KP}^{(h)}\mathbf{e}_{i}^{p}}{\left|W_{KP}^{(h)}\mathbf{e}_{i}^{p}\right|}\right\rangle 
\end{array}
\label{eq:gsa-1}
\end{equation}
where $s_{t+k,\,t+i}^{(h)}$ measures the similarity between the next graph signal $\mathbf{x}_{t+k}$ and the previous graph signal $\mathbf{x}_{t+i}$ under head $h$ (there are $H$ attention heads, $h=1,\ldots,H$); $w, {w_A}, {w_P} \in \mathbb{R}^{+}$ are learnable parameters; $\left|\cdot\right|$ calculates the norm of a vector; $\left\langle \cdot,\:\cdot\right\rangle $
is the inner product between two vectors, measuring their similarity.

The similarity score $s_{t+k,\,t+i}^{(h)}$ comprises three terms:\begin{itemize}
\item First Term (Temporal Neighborhood Comparison): This term assesses the similarity between the next graph signal $\mathbf{x}_{t+k}$ and a previous graph signal $\mathbf{x}_{t+i}$ by analyzing their respective temporal neighborhoods. The temporal neighborhood of a graph signal encompasses the signal itself and the $M - 1$ preceding graph signals, where $M$ denotes the temporal neighborhood's size. This term evaluates multiple graph signals within these neighborhoods, mitigating the effect of individual signal inaccuracies or noise. This approach provides a more robust and accurate characterization of temporal dependency. Matrices $W_{Q}^{(h)}$ and $W_{K}^{(h)}$ in this term are learnable parameters implemented as sparse linear layers.

\item Second Term (Auxiliary Information Comparison): This term estimates similarity by contrasting auxiliary information $\mathbf{e}_{t+k}^{a}$ and $\mathbf{e}_{t+i}^{a}$. Since similar auxiliary information typically influences graph signals in comparable ways, it is crucial to include this aspect in the similarity assessment. Here, $W_{QA}^{(h)}$ and $W_{KA}^{(h)}$
are learnable parameter matrices implemented as fully-connected layers.

\item Third Term (Temporal Positional Encoding Comparison): This term compares the learnable temporal positional encodings $\mathbf{e}_{k}^{p}$ and $\mathbf{e}_{i}^{p}$. Some graph signals exhibit temporal dependencies related to their positions in time (e.g., weekly patterns in taxi demand). The learnable temporal positional encodings capture these dependencies. Matrices $W_{QP}^{(h)}$ and $W_{KP}^{(h)}$, implemented as fully-connected layers, facilitate this comparison.
\end{itemize}

After computing the similarity scores, a softmax layer processes them to derive \textit{attention scores} $\alpha_{t+k,\,t+i}^{(h)}$
as outlined in Equation~\ref{eq:gsa_attention_score}. These scores indicate how much attention is given to each previous graph signal for predicting the next one.
\begin{equation}
\alpha_{t+k,\,t+i}^{(h)}=\frac{\mathrm{exp}\left(s_{t+k,\,t+i}^{(h)}\right)}{\sum_{j=-T+M}^{k-1}{\mathrm{exp}\left(s_{t+k,\,t+j}^{(h)}\right)}}
\label{eq:gsa_attention_score}
\end{equation}

In scenarios where non-stationarity is pronounced and the next graph signal does not resemble any previous ones, the focus shifts to \textit{recent trends.} This is captured by the \textit{extended attention scores} ${\alpha}_{t+k,\,t+i}'^{(h)}$
in Equation~\ref{eq:extended_attention_score}, balancing the focus between capturing temporal dependencies and recent trends.

\begin{equation}
{\alpha}_{t+k,\,t+i}'^{(h)}=\frac{\mathrm{exp}\left(s_{t+k,\,t+i}^{(h)}\right)}{\sum_{j=-T+M}^{k}{\mathrm{exp}\left(s_{t+k,\,t+j}^{(h)}\right)}}
\label{eq:extended_attention_score}
\end{equation}

The primary distinction between Equations~\ref{eq:extended_attention_score} and \ref{eq:gsa_attention_score} lies in the denominator of Equation~\ref{eq:extended_attention_score}, which incorporates the term ${s}_{t+k,\,t+k}^{(h)}$. This term represents the similarity score of the next graph signal with itself. In scenarios where the upcoming graph signal significantly diverges from all previous signals, ${s}_{t+k,\,t+k}^{(h)}$
becomes considerably larger than the other similarity scores. Consequently, the extended attention score ${\alpha}_{t+k,\,t+k}'^{(h)}$ approaches 1, while 
${\alpha}_{t+k,\,t+i}'^{(h)}$ for $i=-T+M,\ldots,k-1$ tends toward 0. This implies an increased focus on the recent trend for making predictions.

We calculate the \textit{refined estimate} $\widehat{\mathbf{e}}_{t+k}$ using Equation~\ref{eq:gsa_decoder_update}. This process begins with the original estimate $\mathbf{e}_{t+k}'$ and involves adding a composite of updates. These updates, represented as $\Delta{\mathbf{e}}_{t+k}'^{(h)}$, are derived from various attention heads, each contributing distinct insights. The formula is:
\begin{equation}
\begin{array}{cll}
\widehat{\mathbf{e}}_{t+k} & \hspace{-0.5em}= & \hspace{-0.5em}\mathbf{e}_{t+k}'+W_{O}\left[\Delta{\mathbf{e}}_{t+k}'^{(1)} \;\Big{\Vert}\; \Delta{\mathbf{e}}_{t+k}'^{(2)}\;\Big{\Vert}\;\cdots\;\Big{\Vert}\;\Delta{\mathbf{e}}_{t+k}'^{(H)}\right]\vspace{1em}\\
\Delta{\mathbf{e}}_{t+k}'^{(h)} &\hspace{-0.5em}  = & \hspace{-0.5em}\sum_{j=-T+1}^{k-1}{\alpha}_{t+k,\,t+j}'^{(h)}\,W_{V}^{(h)}\,\mathbf{e}'_{t+j} + {\alpha}_{t+k,\,t+k}'^{(h)}\mathrm{GRU}\left({\mathbf{e}'_{t+k-M+1},\cdots,\mathbf{e}'_{t+k-1}}\right)
\end{array}
\label{eq:gsa_decoder_update}
\end{equation}

Update $\Delta{\mathbf{e}}_{t+k}'^{(h)}$ encompasses two key components.
\begin{itemize}
    \item The first component uses the extended attention scores ${\alpha}_{t+k,\,t+i}'^{(h)}$ to weight the encoding of each previous graph signal $\mathbf{e}_{t+i}'$, thereby capturing the impact of temporal dependencies. The weighting is facilitated by $W_{V}^{(h)}$, a learnable parameter matrix implemented as a sparse linear layer.
    \item The second component involves GRU~\cite{GRU}, which focuses on capturing the recent trend. It processes the encodings from the recent temporal neighborhood $\{\mathbf{e}_{t+i}'\mid i=k-M+1,\ldots,k-1\}$, predicts the trend, and updates the estimate based on this prediction.
\end{itemize}

Finally, the updates $\Delta{\mathbf{e}}_{t+k}'^{(h)}$ from all attention heads are concatenated and processed through the learnable matrix $W_{O}$ to form a composite update. This composite, once added to the original estimate $\mathbf{e}_{t+k}'$, gives us the refined estimate $\widehat{\mathbf{e}}_{t+k}$. This refined estimate thus reflects an integration of insights from both the historical temporal dependencies and the recent trends.

\subsection{\label{sec:gsa_encoding}Graph Sequence Attention (Filtering)}
The aim of GSA (filtering) within each encoder layer is to diminish noise in the historical graph signal encodings. As shown in Figure~\ref{fig:GSA_Forecaster}(b), GSA (filtering) takes two key inputs:
\begin{itemize}
\item Entry A: Encodings of historical graph signals $\{\mathbf{e}_{t+i}\mid i=-T+1,\ldots,0\}$
\item Entry B: Auxiliary information  $\left\{\mathbf{e}_{t+i}^a\mid i=-T+1,\ldots,0\right\}$
\end{itemize}
It outputs the noise-reduced encodings $\{\mathbf{e}_{t+i}'\mid i=-T+1,\ldots,0\}$. The process to achieve these filtered encodings is described as follows.

Initially, GSA (filtering) evaluates the similarity scores $s_{t+i,\,t+j}^{(h)}$ between any two graph signals, $\mathbf{x}_{t+i}$ and $\mathbf{x}_{t+j}$. These scores are pivotal in identifying graph signals with close true values. By aggregating similar signals, we can mitigate their noise according to the central limit theorem~\cite{wasserman2013all}. The similarity scores are calculated using Equation~\ref{eq:gsa-filter}:

\begin{equation}
\begin{array}{lll}
\hspace{-0.5em}s_{t+i,\,t+j}^{(h)} & \hspace{-0.7em}= &\hspace{-0.7em} \dfrac{1}{l_2-l_1+1}{\mathlarger{\sum}\limits_{m = -M_1}^{M_2} \hspace{-0.5em} w\left\langle {\frac{W_{Q}^{(h)}\mathbf{e}_{t+i+m}}{\left|W_{Q}^{(h)}\mathbf{e}_{t+i+m}\right|}},\;{\frac{W_{K}^{(h)}\mathbf{e}_{t+j+m}}{\left|W_{K}^{(h)}\mathbf{e}_{t+j+m}\right|}}\right\rangle}\vspace{0.5em} + w_{A}\left\langle \frac{W_{QA}^{(h)}\mathbf{e}_{t+i}^{a}}{\left|W_{QA}^{(h)}\mathbf{e}_{t+i}^{a}\right|},\;\frac{W_{KA}^{(h)}\mathbf{e}_{t+j}^{a}}{\left|W_{KA}^{(h)}\mathbf{e}_{t+j}^{a}\right|}\right\rangle + w_{P}\left\langle \frac{W_{QP}^{(h)}\mathbf{e}_{i}^{p}}{\left|W_{QP}^{(h)}\mathbf{e}_{i}^{p}\right|},\;\frac{W_{KP}^{(h)}\mathbf{e}_{j}^{p}}{\left|W_{KP}^{(h)}\mathbf{e}_{j}^{p}\right|}\right\rangle 
\end{array}
\label{eq:gsa-filter}
\end{equation}
The similarity score comprises three components. The first component gauges similarity by comparing temporal neighborhoods, which include the graph signal itself, $M_1$ preceding signals, and $M_2$ subsequent signals, with certain exceptions at the start and end of the sequence. This approach ensures that the similarity scores are less influenced by signal noise. The second component is based on auxiliary information, while the third component assesses similarity using learnable temporal positional encodings 
$\{\mathbf{e}_{i}^{p}\mid i=-T+1,\ldots,0\}$.
The parameter matrices $W_{Q}^{(h)}$ and $W_{K}^{(h)}$ are in sparse linear layers, whereas $W_{QA}^{(h)}$, $W_{KA}^{(h)}$,  $W_{QP}^{(h)}$, and $W_{KP}^{(h)}$ are in dense layers. Parameters $w$, $w_A$, and $w_P$ are also learnable, enhancing the model's adaptability in similarity assessment.

After computing these scores, GSA (filtering) employs a softmax layer to derive the attention scores $\alpha_{t+i,\,t+j}^{(h)}$ as per Equation~\ref{eq:attention_score}. These scores, ranging between 0 and 1, reflect the attention allocated to each graph signal $\mathbf{x}_{t+j}$ for filtering $\mathbf{x}_{t+i}$.
\begin{equation}
\alpha_{t+i,\,t+j}^{(h)}=\frac{\mathrm{exp}\left(s_{t+i,\,t+j}^{(h)}\right)}{\sum_{n=-T+1}^{0}{\mathrm{exp}\left(s_{t+i,\,t+n}^{(h)}\right)}}
\label{eq:attention_score}
\end{equation}

Finally, GSA (filtering) uses these attention scores as weights to combine graph signals with similar true values, thereby filtering out noise. The filtered encoding $\mathbf{e}_{t+i}'$ for each graph signal $i=-T+1,\ldots,0$
is calculated using Equation~\ref{eq:filter_update}:
\begin{equation}
\begin{array}{cll}
\mathbf{e}_{t+i}' & \hspace{-0.5em}= & \hspace{-0.5em}\mathbf{e}_{t+i}+W_{O}\left[\Delta{\mathbf{e}}_{t+i}^{(1)} \;\Big{\Vert}\; \Delta{\mathbf{e}}_{t+i}^{(2)}\;\Big{\Vert}\;\cdots\;\Big{\Vert}\;\Delta{\mathbf{e}}_{t+i}^{(H)}\right]\vspace{1em}\\
\Delta{\mathbf{e}}_{t+i}^{(h)} &\hspace{-0.5em}  = & \hspace{-0.5em}\sum_{j=-T+1}^{0}{\alpha}_{t+i,\,t+j}^{(h)}\,W_{V}^{(h)}\,\mathbf{e}_{t+j}
\end{array}
\label{eq:filter_update}
\end{equation}
where $\Delta{\mathbf{e}}_{t+i}^{(h)}$ represents the update to the original encoding $\mathbf{e}_{t+i}$ under each head $h$. Both $W_O$ and $W_{V}^{(h)}$
are learnable parameter matrices, implemented as sparse linear layers.

\subsection{Optional Extension: Node-Level Attention \label{sec:node_level_attention}}
The GSA-Forecaster primarily employs a single scalar to represent the similarity between two graph signals, as detailed in Equations~\ref{eq:gsa-1} and \ref{eq:gsa-filter}. This approach, where different nodes share the same attention score, is termed graph-level attention. Essentially, the entire graph is assigned a uniform attention score. However, this method may not be optimal for datasets characterized by strong non-stationarity. For instance, unforeseen events might temporarily affect the data related to a specific node, leading to a situation where that node's attention score differs from others. To address this, we propose an optional node-level attention mechanism. This enhancement allows for individual nodes to have distinct attention scores, effectively handling non-stationarity.

\begin{figure}[th]
\centering
\includegraphics[width=1.0\columnwidth]{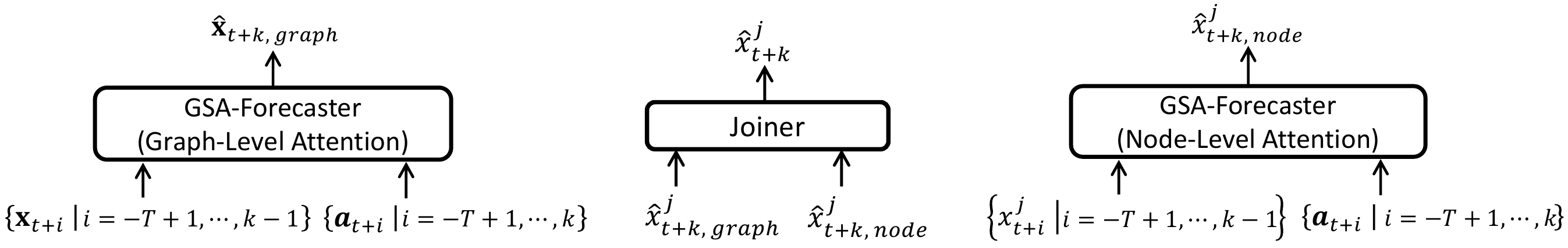}
\caption{Extending GSA-Forecaster with node-level attention.}
\label{fig:joiner}
\end{figure}

Figure~\ref{fig:joiner} demonstrates our extension of GSA-Forecaster, which now includes two variants: GSA-Forecaster (Graph-Level Attention) and GSA-Forecaster (Node-Level Attention).

The GSA-Forecaster (Graph-Level Attention), as described in Sections~\ref{sec:arch_gsa} to \ref{sec:gsa_encoding}, processes a series of graph signals $\{\mathbf{x}_{t+i}\mid i=-T+1,\ldots,k-1\}$ and auxiliary information $\{\mathbf{a}_{t+i}\mid i=-T+1,\ldots,k\}$ to predict the next graph signal $\widehat{\mathbf{x}}_{t+k, \textrm{graph}}$. A graph signal here is a vector comprising data from each node. For example, the predicted next graph signal $\widehat{\mathbf{x}}_{t+k, \textrm{graph}}$ is an array $\left[{\widehat{x}_{t+k, \textrm{graph}}^{1}}, \cdots, {\widehat{x}_{t+k, \textrm{graph}}^{N}}\right]$ where $\widehat{x}_{t+k,\textrm{graph}}^{j}$ represents the prediction for node $j$. 

In contrast, GSA-Forecaster (Node-Level Attention) is tailored for individual node predictions. It mirrors the structure discussed in Sections~\ref{sec:arch_gsa} to \ref{sec:gsa_encoding} but with a key alteration: it predicts separately for each node.
For a given node $j$, the model exclusively processes the historical data of that node, $\{x_{t+i}^{j}\mid i=-T+1,\ldots,k-1\}$, combined with the auxiliary information, to make its prediction $\widehat{x}_{t+k, \textrm{node}}^{j}$.
This version modifies the architecture by replacing sparse linear layers with dense layers, enabling distinct attention scores for each node, particularly useful during non-stationary conditions. However, it omits the use of temporal positional encodings, which are more relevant during stationary periods (setting $w_P = 0$ in Equations~\ref{eq:gsa-1} and \ref{eq:gsa-filter}).

The final predictions from both GSA-Forecasters ($\widehat{x}_{t+k, \textrm{graph}}^{j}$ and $\widehat{x}_{t+k, \textrm{node}}^{j}$)
are combined by the joiner module to produce the conclusive prediction 
$\widehat{x}_{t+k}^{j}$. While the joiner could employ complex strategies for this combination, we opt for a simpler approach:
\begin{equation}
\widehat{x}_{t+k}^{j} = \gamma\times\widehat{x}_{t+k, \textrm{graph}}^{j} + (1-\gamma)\times\widehat{x}_{t+k, \textrm{node}}^{j}   
\end{equation}
Here, $\gamma$ is a hyperparameter determined based on the validation set. This process is repeated for each node's data prediction.

Node-level attention is an optional extension used exclusively for datasets exhibiting significant non-stationarity. In our evaluations, we apply this extension specifically to the PEMS-BAY dataset.

\subsection{Theoretical Analysis: Employing GRU to Learn From the Recent Trend}

When nonstationarity occurs, historical temporal dependencies become unreliable for predicting future graph signals. To address this, GSA-Forecaster employs a GRU to predict future graph signals based on the recent trend. This subsection provides a theoretical analysis of how GSA-Forecaster achieves this. We begin with the following theorem:
\begin{theorem}
When predicting the $({t+k})^\mathrm{th}$ graph signal, if the similarity scores satisfy the following condition: 
\[
    s_{t+k,\,t+k}^{(h)} = \max\left\{s_{t+k,\,t-T+M,}^{(h)}\quad s_{t+k,\,t-T+M+1,}^{(h)}\quad \dots\quad s_{t+k,\,t+k-1}^{(h)}\right\} + \delta s
\]
then the extended attention score ${\alpha}_{t+k,\,t+k}'^{(h)}$ (representing the attention given to the recent trend) satisfies:
\[
    \alpha_{t+k,\,t+k}'^{(h)} \geq \frac{1}{1 + (k + T - M) \exp(- \delta s)}
\]
where $T$ is the number of historical graph signals and $M$ denotes the size of the temporal neighborhood.
\end{theorem}

\begin{proof}
From the definition of the extended attention score shown in Eq.~\eqref{eq:extended_attention_score}:
\[
\alpha_{t+k,\,t+k}'^{(h)} 
= \frac{\exp\left(s_{t+k,\,t+k}^{(h)}\right)}{\sum_{j=-T+M}^{k} \exp\left(s_{t+k,\,t+j}^{(h)}\right)}
\]
Rewriting the numerator and the denominator:

\[
\alpha_{t+k,\,t+k}'^{(h)} 
= \frac{1}{\sum_{j=-T+M}^{k} \exp\left(s_{t+k,\,t+j}^{(h)} - s_{t+k,\,t+k}^{(h)}\right)}
\]
Given the condition on the similarity scores, we can bound the denominator:
\[
\alpha_{t+k,\,t+k}'^{(h)} 
\geq \frac{1}{1 + \sum_{j=-T+M}^{k-1} \exp\left(- \delta s\right)}
\]
Simplifying further:
\[
\alpha_{t+k,\,t+k}'^{(h)} 
\geq \frac{1}{1 + (k + T - M) \exp(- \delta s)}
\]
\end{proof}

\paragraph{Analysis}
In cases of nonstationarity, historical temporal dependencies are minimal, and the similarity scores behave as:
\[
s_{t+k,\,t+k}^{(h)} - s_{t+k,\,t+j}^{(h)} \approx w, \quad j = -T+M, \cdots, k-1
\]
where $w$ is the learnable weight used to compute the similarity scores in Eq.~\eqref{eq:gsa-1}.

Thus, $\delta s \approx w$, and the term $\left(k + T - M\right) \exp\left(- \delta s\right) \approx 0$. Consequently, the extended attention score for the recent trend approaches:
\[
\alpha_{t+k,\,t+k}'^{(h)} \approx 1
\]
For instance, in our model trained on the NYC Taxi dataset, $k + T - M < 300$ and $w = 10 \text{ to } 20$, the term $\left(k + T - M\right) \exp(- \delta s)$ becomes negligible, $\leq 0.02$. This implies $\alpha_{t+k,\,t+k}'^{(h)} \approx 1$, while attention given to historical signals diminishes (because the sum of the extended attention scores is 1):
\[
\alpha_{t+k,\,t+j}'^{(h)} \approx 0, \quad j = -T+M, \cdots, k-1
\]

Based on Eq.~\eqref{eq:gsa_decoder_update}, under such scenarios, the refined embedding for the predicted $({t+k})^\mathrm{th}$ graph signal is updated as:
\[
\Delta{\mathbf{e}}_{t+k}'^{(h)} \approx \mathrm{GRU}\left(\mathbf{e}'_{t+k-M+1}, \cdots, \mathbf{e}'_{t+k-1}\right)
\]
Thus, the GRU plays a central role in capturing the recent trend, effectively driving the prediction.

\subsection{Computational Complexity and Scalability}
Similar to other Transformer-based models, the major computational complexity of GSA-Forecaster lies in the attention module. The attention module's computation involves the calculation of attention scores and output embeddings. Let $N$ be the number of locations, $d_\textrm{model}$ the embedding dimension, $T$ the time length of the historical data, and $T'$ the time length of the prediction. For models based on standard attention, the complexity of calculating both attention scores and output embeddings is $O(Nd_\textrm{model}(T^2 + TT' + \frac{1}{2} T'^2))$. For GSA-Forecaster, the complexity of calculating embeddings remains the same, while the complexity of attention score calculation is $O(Nd_\textrm{model}(T^2 + TT' + \frac{1}{2} T'^2) + M(T+T'))$, where $M$ represents the temporal neighborhood size. From this analysis, we can see that the additional complexity introduced by GSA-Forecaster over standard attention-based models is minimal, as the temporal neighborhood size $M$ is generally much smaller than the time length of the historical and predicted data and the number of locations. Similar to standard attention-based models, the complexity of GSA-Forecaster grows linearly with the number of locations and quadratically with the time length of the historical and predicted data.

We evaluate the execution time of graph sequence attention by applying it to forecast the NYC Taxi dataset, comparing it to the execution time of standard attention. Both models are executed on a single NVIDIA GeForce RTX 2080 Ti GPU. Table~\ref{tab:inference} reports the execution time of graph sequence attention across different temporal neighborhood sizes. As observed, graph sequence attention demonstrates execution times comparable to standard attention, with minimal overhead introduced by incorporating the temporal neighborhood.

\begin{table}
\centering
\caption{Execution time comparison between graph sequence attention and standard attention for forecasting NYC Taxi demand.}
\renewcommand{\arraystretch}{1.2}
\resizebox{\textwidth}{!}{%
\begin{tabular}{|c|c|c|c|}
\hline
 & \textrm{Standard Attention} & \makecell{\textrm{Graph Sequence Attention} \\ \textrm{(Temporal Neighborhood Size = 10)}} & \makecell{\textrm{Graph Sequence Attention} \\ \textrm{(Temporal Neighborhood Size = 20)}} \\
\hline
\textrm{Time (millisecond/sample)} & \makecell{10.264 $\pm$ 0.414} & \makecell{10.327 $\pm$ 0.405} & \makecell{10.391 $\pm$ 0.364} \\
\hline
\end{tabular}
}
\label{tab:inference}
\end{table}

\section{\label{sec:evaluation}Evaluation Setups}
In this section, we detail our methodology for assessing the forecasting accuracy of GSA-Forecaster.

\subsection{Datasets}
\subsubsection{NYC Taxi}\hfill

\textbf{\textit{Hourly Taxi Demand.}} We utilized the dataset provided by the New York City Taxi and Limousine Commission~\cite{NYC_Taxi}, covering the period from January 1, 2009, to June 30, 2016. This comprehensive dataset includes information on approximately 1.063 billion taxi rides in Manhattan over 7.5 years. We focused on 471 specific locations in Manhattan, tracking the hourly taxi demand at each.\footnote{To determine the key locations in Manhattan for measuring hourly taxi demand from a roadmap comprising 5,464 locations, we applied two criteria: a geometric threshold and a demand threshold. Specifically, we selected locations based on a minimum distance requirement of 150 meters between any two locations (geometric threshold) and a minimum average hourly taxi demand of 20 pickups (demand threshold). The average hourly demand at each selected location was calculated by mapping the total taxi pickups in Manhattan to these chosen locations only. By adhering to these thresholds, we narrowed down our focus to 471 strategic locations.} The `hourly taxi demand' represents the number of taxi pickups occurring near each location every hour. Following Forecaster~\cite{Forecaster}, we used the Gaussian Markov random fields theory to construct a graph capturing the spatial dependencies among these locations. This way, we developed a large-scale, graph-based, time-dependent dataset that encompasses over 31 million data points (derived from 471 nodes and the hourly data collected over 7.5 years).

\textbf{\textit{Auxiliary Information.}}
To augment our analysis, we incorporated hourly weather data for Manhattan sourced from Weather Underground~\cite{WeatherUnderground}. This data, spanning the same period as the taxi dataset, includes metrics such as temperature, precipitation, visibility, wind speed, and indicators for snow, rain, and fog.

\textbf{\textit{Data Split.}} Following the setup in prior work~\cite{Forecaster}, we divided the hourly taxi demand and auxiliary weather data into three distinct sets:
\begin{itemize}[leftmargin=15pt]
    \item Training set (80\% of the data): 01/01/2009 -- 12/31/2011 and 07/01/2012 -- 06/30/2015 
\item Validation set (6.7\% of the data): 01/01/2012 -- 06/30/2012  
\item Test set (13.3\% of the data): 07/01/2015 -- 06/30/2016 
\end{itemize}

\subsubsection{PEMS-BAY}\hfill

\textbf{\textit{Traffic Speed}}.
The PEMS-BAY dataset~\cite{DCRNN_ICLR2018} compiles traffic speed data from 325 sensors across the San Francisco Bay Area. This dataset spans six months, from January 1, 2017, to June 30, 2017. Traffic speed readings are recorded in five-minute intervals, resulting in approximately 17 million data points (calculated from 325 nodes and the total number of five-minute periods across six months). In line with Forecaster~\cite{Forecaster}, we utilized the Gaussian Markov random fields theory to establish a spatial dependency graph.

\textbf{\textit{Auxiliary Information}}. Due to challenges in acquiring high-quality weather data for the sensor locations, we did not incorporate auxiliary information in this dataset.

\textbf{\textit{Data Split}}. The dataset was divided as follows: the training set comprises the initial 70\% of the data, followed by the validation set with 10\%, and the test set encompassing the final 20\%. This setup is the same as prior work~\cite{DCRNN_ICLR2018}.

\subsubsection{ECL}\hfill

\textbf{\textit{Electricity Consumption Load}}.
The ECL dataset~\cite{DCRNN_ICLR2018} contains hourly electricity consumption loads for 321 clients over a two-year period. Consistent with the approach used in Forecaster~\cite{Forecaster}, we applied Gaussian Markov random fields theory to construct a graph for this data.

\textbf{\textit{Data Split}}.
Following previous work~\cite{informer_ecl}, we divided the dataset into training, validation, and test sets, corresponding to 15, 3, and 4 months of data, respectively.

\subsubsection{Traffic}\hfill

\textbf{\textit{Road Occupancy Rate}}.
The Traffic dataset~\cite{DCRNN_ICLR2018} includes hourly road occupancy rates at 860 locations over a two-year period. Similar to our approach with the ECL dataset, we used Gaussian Markov random fields theory to create a graph for this data, following the methodology in Forecaster~\cite{Forecaster}.

\textbf{\textit{Data Split}}.
Following prior work~\cite{autoformer_traffic}, we partitioned the dataset into training, validation, and test sets, comprising 70\%, 10\%, and 20\% of the data, respectively.
\subsection{Forecasting Tasks}
\subsubsection{NYC Taxi}
In line with previous studies \cite{Forecaster}, we aim to forecast the next three hours of taxi demand in New York City based on historical taxi demand data from the preceding month and relevant auxiliary information.
Specifically, letting the current time be $t$ and $\mathbf{x}_{t}$ be hourly taxi demand, we use the following historical data (together with auxiliary information):
\begin{itemize}
\item The past week: $\mathbf{{x}}_{t+i-j\times24}$, $i=-23,\ldots,0,\,j=0,\ldots,6$
\item Relevant hours on the same weekday of the past four weeks: $\mathbf{{x}}_{t+i-j\times24\times7}$, $\{i=-18,\ldots,5,\, j=2,\ldots,4\}$\\ and $\{i=-18,\ldots,0,\,j=1\}$
\end{itemize}

\subsubsection{PEMS-BAY}
Our objective is to forecast traffic speed for every five-minute interval in the upcoming hour based on data from similar intervals in the previous month.
Specifically, letting the current time be $t$ and $\mathbf{x}_{t}$ denote traffic speed, we incorporate the following data in our prediction:\begin{itemize}
\item The past hour:
$\mathbf{{x}}_{t+i}$, $i=-11,\ldots,0$
\item Relevant intervals in the past week: $\mathbf{{x}}_{t+i-j\times24\times12}$, $i=-11,\ldots,18,\,j=1,\ldots,6$
\item Relevant intervals on the same weekday of the past four weeks: $\mathbf{{x}}_{t+i-j\times7\times24\times12}$, $i=-11,\ldots,18,\,j=1,\ldots,4$
\end{itemize}

\subsubsection{ECL and Traffic}
The goal is to forecast hourly data for the next 48 hours (for ECL) and 24 hours (for Traffic) using data from similar intervals in the previous month. Specifically, let the current time be $t$ and let $\mathbf{x}_{t}$ represent traffic speed. The following data is incorporated into our prediction:

\begin{itemize}
\item Relevant intervals from the past eight days: $\mathbf{{x}}_{t-i}$, where $i=0,\ldots,8\times24-1$
\item Relevant intervals on the same weekday from the past four weeks: $\mathbf{{x}}_{t+i-j\times7\times24}$, where $i=-24,\ldots,48$ and $j=2,\ldots,4$
\end{itemize}

\subsection{Metrics}
Consistent with previous studies \cite{DCRNN_ICLR2018, STMGCN_AAAI2019,Forecaster}, our evaluation uses the root-mean-square error (RMSE) \cite{metrics} and the mean absolute percentage error (MAPE) \cite{metrics} to measure the accuracy of the forecasting results on NYC Taxi and PEMS-BAY. RMSE characterizes the absolute error, while MAPE characterizes the relative error. Lower RMSE and MAPE mean better accuracy.

For the NYC Taxi dataset, following the practice in prior work~\cite{STMGCN_AAAI2019,  Forecaster}, we set a threshold on taxi demand when calculating MAPE: if $x_t^i < 20$ ($x_t^i$ refers to the hourly taxi demand at location $i$ at time $t$), exclude $x_t^i$ from the calculation of MAPE.

In line with previous studies~\cite{crossformer, informer_ecl, autoformer_traffic}, we evaluate the accuracy of the forecasting results on ECL and Traffic using mean squared error (MSE) and mean absolute error (MAE).

\subsection{Implementation Details} 
GSA-Forecaster uses the following loss function:
\begin{equation}
\textrm{{loss}}\left(\cdot\right)=\eta\times\textrm{{RMSE}}^2+\textrm{{MAPE}}
\end{equation}
where a constant weight $\eta$ is used
such that $\textrm{RMSE}$ and $\textrm{MAPE}$ have almost equal contributions to the loss function.

\section{\label{sec:evaluation_results}Evaluation Results}
In this section, we present the results of our evaluation, focusing on the forecasting accuracy of GSA-Forecaster and its comparison with leading predictive models.

\subsection{Comparison on Synthetic Data}

\begin{figure}[th]
\centering
\includegraphics[width=0.6\columnwidth]{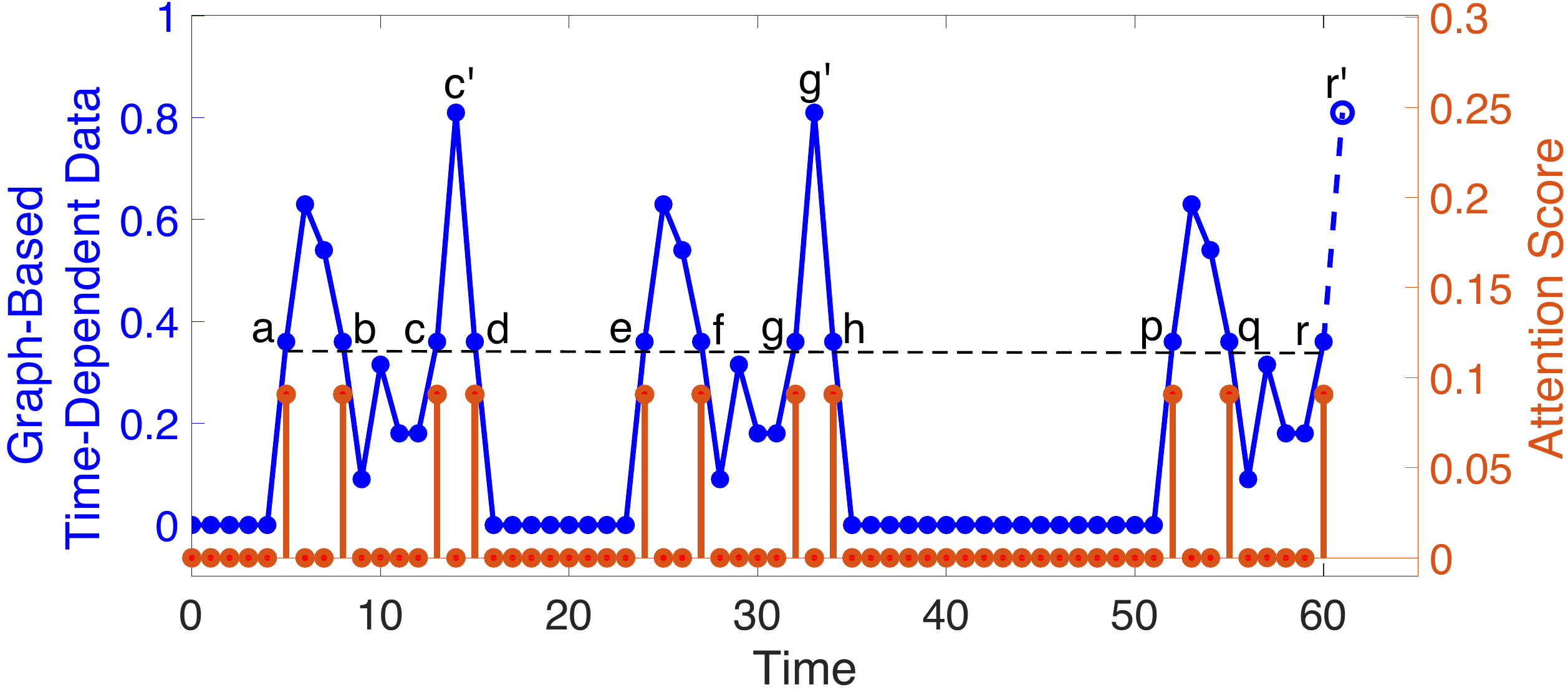}
\caption{Synthetic dataset and corresponding attention scores calculated using standard attention.}
\label{fig:standard_attention}
\end{figure}

Our first experiment contrasts graph sequence attention with standard attention using synthetic data to highlight their differences. Figure~\ref{fig:standard_attention} presents a graph-based, time-dependent dataset (illustrated by a blue curve) consisting of a single node. This dataset displays a repetitive pattern, though it is not periodic. The data at each time point, $x_t$, is encoded into 
$\mathbf{e}_t = \left[e_{t, 1}, \cdots, e_{t, n}\right]^{\top}$. The encoding formula is as follows:
\begin{equation}
    {x_t} = \sum_{i=1}^{n}{\left(\frac{e_{t, i}+1}{2}\right)\cdot{2^{-i}}}
    \label{standard_encoding}
\end{equation}
where $n=8$, $e_{t, i} \in \{1,-1\}$. For simplicity, both standard and graph sequence attention in our experiment use a single attention head, identity parameter matrices (i.e., $W_Q = W_K = I$), and exclude auxiliary information or temporal positional encodings. They generate \textit{initial} predictions for the next data point ($r'$) based on the current data point ($r$). 

\begin{figure}[h]
\centering
\includegraphics[width=0.6\columnwidth]{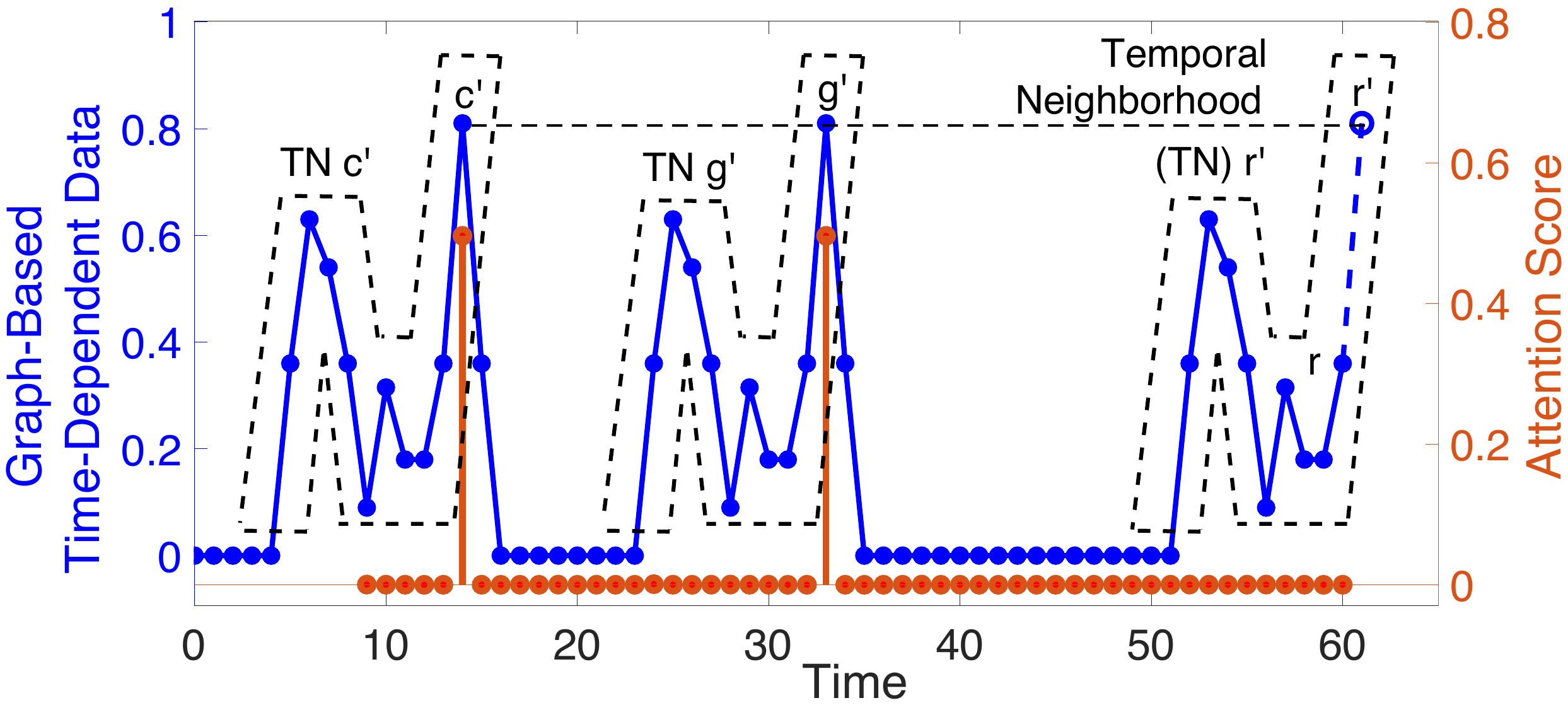}
\caption{Synthetic dataset and corresponding attention scores calculated using graph sequence attention.}
\label{fig:GSA_Result}
\end{figure}

Figure~\ref{fig:standard_attention} illustrates how standard attention predicts the next data point ($r'$). It shows that standard attention focuses on historical data points ($a$, $b$, $c$, ..., $r$) resembling the current data ($r$) rather than those ($c'$ and $g'$) resembling the next data point. This occurs because standard attention only compares the initial estimate, which is derived from the current data and contains errors, with historical data points. As a result, this leads to an erroneous prediction.

In contrast, Figure~\ref{fig:GSA_Result} displays the attention scores using graph sequence attention for predicting the next data point ($r'$). This method examines the temporal neighborhood of the next data and compares it to that of historical data points. Consequently, it avoids the pitfalls of errors in individual data points, such as those in the initial estimate. As depicted, graph sequence attention accurately identifies historical data points ($c'$ and $g'$) similar to the next data point, effectively capturing the temporal dependency.

\subsection{Performance on NYC Taxi}
Table~\ref{table:forecasting_accuracy} showcases a comparison of GSA-Forecaster with baseline models (VAR~\cite{VAR}, DCRNN~\cite{DCRNN_ICLR2018}, Transformer~\cite{Transformer}, GMAN~\cite{GMAN}, AGCRN~\cite{AGCRN}, Graph WaveNet~\cite{GraphWaveNet}, Forecaster~\cite{Forecaster}, and Crossformer~\cite{crossformer}) in forecasting hourly taxi demand in New York City. It assesses overall accuracy and the accuracy for each of the upcoming three hours. The table indicates that among the baseline models, Forecaster excels over DCRNN, Graph WaveNet, GMAN, AGCRN, and Transformer. Our GSA-Forecaster outperforms all, including Forecaster and Crossformer, by enhancing the capability to capture temporal dependencies through graph sequence attention. Against Forecaster, the most competitive baseline, GSA-Forecaster shows superior predictions for each of the three future hours, reducing overall RMSE and MAPE by 6.7\% and 5.8\%, respectively.

\begin{table*}[th]
\caption{\label{table:forecasting_accuracy}Accuracy of GSA-Forecaster
and baseline models on forecasting hourly taxi demand in New York City.}
\centering
\begin{tabu}{|c|c|c|c|c|c|c|c|c|}
\hline 
\multirow{2}{*}{Model} & \multicolumn{2}{c|}{Overall} & \multicolumn{2}{c|}{Next Hour} & \multicolumn{2}{c|}{Second Next Hour} & \multicolumn{2}{c|}{Third Next Hour}\tabularnewline
\cline{2-9} 
 & RMSE & MAPE & RMSE & MAPE & RMSE & MAPE & RMSE & MAPE \tabularnewline
\hline 
VAR & 12.57 & 33.21\% & 11.10 & 30.25\% & 12.86 & 33.83\% & 13.63 & 35.55\%\tabularnewline
\hline
DCRNN & 8.74 & 19.83\% & 8.29 & 19.05\% & 8.81 & 19.91\% & 9.11 & 20.52\%\tabularnewline
\hline
Transformer & 8.87 & 17.86\% & 8.43 & 17.01\% & 8.89 & 17.88\% & 9.25 & 18.70\%\tabularnewline
\hline
GMAN &  10.81 & 23.47\% & 10.78 & 23.45\% & 10.77 & 23.33\% & 10.88 & 23.60\% \tabularnewline
\hline
AGCRN &  8.60 & 19.60\% & 8.25 & 19.03\% & 8.65 & 19.61\% & 8.88 & 20.15\% \tabularnewline
\hline 
Graph WaveNet &  8.86 & 20.30\% & 8.46 & 19.79\%  & 8.88 & 20.27\% & 9.23 & 20.81\% \tabularnewline
\hline 
Crossformer &  8.69 & 20.99\% & 8.32 & 20.33\%  & 8.77 & 21.08\% & 8.96 & 21.55\% \tabularnewline
\hline 
Forecaster & 8.41 & 16.56\% & 7.89 & 15.93\% & 8.48 & 16.62\% & 8.83 & 17.14\% \tabularnewline
\hline
\makecell{GSA-Forecaster \\(no auxiliary information)} & 7.92 & 15.76\%  & 7.62 & 15.38\% & 7.98 & 15.82\% & 8.15 & 16.07\% 
\tabularnewline 
\hline 
GSA-Forecaster & \textbf{7.85} & \textbf{15.60\%} & \textbf{7.59} & \textbf{15.28\%} & \textbf{7.90} & \textbf{15.66\%} & \textbf{8.05} & \textbf{15.86\%} \tabularnewline
\hline 
\end{tabu}
\end{table*}

An additional analysis was conducted to assess the impact of auxiliary information on GSA-Forecaster's performance. Results indicate that while auxiliary information does enhance forecasting accuracy, GSA-Forecaster maintains a significant lead over baseline models even without it.

\begin{figure*}[ht]
\centering
\includegraphics[width=1.0\columnwidth]{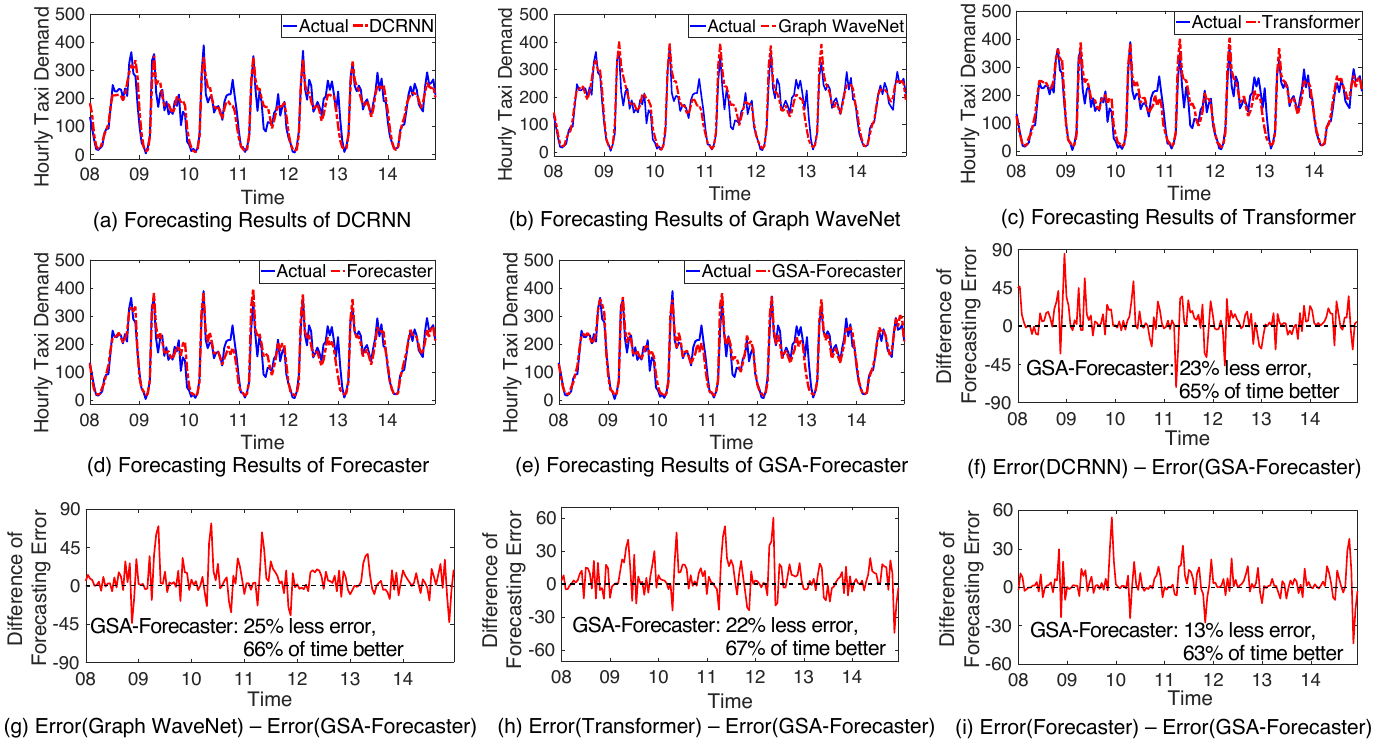}
\caption{Three-hour-ahead forecasting for the taxi demand near New York Penn Station from Sunday, November 8, 2015 at 12:00 a.m.~to Saturday, November 14, 2015 at 11:59 p.m. Error (model) represents the forecasting error of a model, i.e., the absolute value of the difference between its predicted taxi demand and the actual taxi demand. Each tick label ``DD'' in the x-axis refers to the beginning of a date (e.g., ``08'' means November 8, 2015 at 12:00 a.m.).}
\label{fig:model_trace}
\end{figure*}

For a tangible comparison, Figure~\ref{fig:model_trace} (a) – (e) display the three-hour-ahead forecasting accuracy of different models around New York Penn Station for the week of November 8 – 14, 2015. Figures~\ref{fig:model_trace} (f) – (i) contrast the forecasting errors of DCRNN, Graph WaveNet, Transformer, and Forecaster with GSA-Forecaster, with a positive difference indicating a larger error in the baseline models. GSA-Forecaster exhibits superior accuracy overall. Specifically, it demonstrates a smaller error than Forecaster 63\% of the time and, on average, shows 13\% less error.

\subsection{Performance on PEMS-BAY}
In Table~\ref{table:forecasting_accuracy_secondary}, the performance of GSA-Forecaster is compared against state-of-the-art models in forecasting traffic speed in the Bay Area. This comparison covers predictions for short-term (15 minutes), mid-term (30 minutes), and long-term (60 minutes) intervals. GSA-Forecaster demonstrates superior performance over models including DCRNN, Transformer, Forecaster, GMAN, and AGCRN across all categories. Notably, when compared to the most effective baseline model, Graph WaveNet, GSA-Forecaster achieves a reduction in RMSE and MAPE for long-term forecasts (60 minutes ahead) by 9.1\% and 5.8\%, respectively. This enhanced performance, particularly in long-term forecasting, is attributed to GSA-Forecaster's advanced attention mechanism which effectively captures long-term temporal dependencies.

\begin{table}[h]
\caption{\label{table:forecasting_accuracy_secondary}Accuracy of GSA-Forecaster and baseline models on forecasting traffic speed in the San Francisco Bay Area.}
\centering
\begin{tabu}{|c|c|c|c|c|c|c|c|c|c|c|}
\hline
  \multirow{2}{*}{Model} & \multicolumn{2}{c|}{15 minutes} & \multicolumn{2}{c|}{30 minutes} & \multicolumn{2}{c|}{60 minutes}\tabularnewline
\cline{2-7} 
  &  RMSE & MAPE & RMSE & MAPE & RMSE & MAPE\tabularnewline
\hline
VAR &  3.16 & 3.60\% & 4.25 & 5.00\% & 5.44 & 6.50\% \tabularnewline\cline{1-7} 
 DCRNN &  2.95 & 2.90\% & 3.97 & 3.90\% & 4.74 & 4.90\% \tabularnewline\cline{1-7} 
 Transformer & 3.98 & 4.83\% & 4.87 & 5.87\% & 5.78 & 6.86\%  \tabularnewline\cline{1-7}
 Forecaster & 3.07 & 3.34\% & 4.26 & 4.51\% & 5.40 & 5.69\%  \tabularnewline\cline{1-7}
 GMAN & 2.90 & 2.91\% & 3.78 & 3.78\% & 4.39 & 4.52\% \tabularnewline\cline{1-7}
 AGCRN & 2.87 & 2.92\% & 3.84 & 3.89\% & 4.72 & 4.90\% \tabularnewline\cline{1-7}
 Graph WaveNet & 2.74 & 2.73\% & 3.70 & 3.67\% & 4.52 & 4.63\% \tabularnewline\cline{1-7}
 \makecell{GSA-Forecaster \\(no node-level attention)} & 2.97 & 3.19\% & 3.71 & 3.93\% & 4.29 & 4.60\% \tabularnewline\cline{1-7}
 \tabulinesep=1.5mm  GSA-Forecaster  & \textbf{2.73} & \textbf{2.72\%} & \textbf{3.54} & \textbf{3.65\%} & \textbf{4.11} & \textbf{4.36\%} \tabularnewline\cline{1-7}
\hline 
\end{tabu}
\end{table}

\subsection{Performance on ECL and Traffic}
Table~\ref{table:forecasting_accuracy_ecl_traffic} presents a comparison of GSA-Forecaster's performance with that of baseline models (LSTMa~\cite{lstma}, LSTnet~\cite{LSTnet}, MTGNN~\cite{mtgnn}, Transformer~\cite{Transformer}, Informer~\cite{informer_ecl}, Autoformer~\cite{autoformer_traffic}, Pyraformer~\cite{pyraformer}, FEDformer~\cite{fedformer}, and Crossformer~\cite{crossformer}) for 48-hour-ahead forecasting on the ECL dataset and 24-hour-ahead forecasting on the Traffic dataset. The baseline results are sourced from the Crossformer paper~\cite{crossformer}. The results demonstrate that GSA-Forecaster consistently outperforms the baseline models across both ECL and traffic datasets.

\begin{table}[h]
\caption{\label{table:forecasting_accuracy_ecl_traffic}Comparison of GSA-Forecaster and Baseline Models in 48-Hour-Ahead Forecasting for Electricity Consumption Loads (ECL) and 24-Hour-Ahead Forecasting for Road Occupancy Rates (Traffic).}
\centering
\begin{tabu}{|c|c|c|c|c|c|c|c|c|}
\hline
  \multirow{2}{*}{Model} & \multicolumn{2}{c|}{ECL} & \multicolumn{2}{c|}{Traffic}\tabularnewline
\cline{2-5} 
  &  MSE & MAE & MSE & MAE \tabularnewline
\hline
LSTMa &  0.486 & 0.572 & 0.668 & 0.378 \tabularnewline\cline{1-5} 
LSTnet &  0.369 & 0.445 & 0.648 & 0.403 \tabularnewline\cline{1-5} 
MTGNN &  0.173 & 0.280 & 0.506 & 0.278 \tabularnewline\cline{1-5} 
 Transformer &  0.334 & 0.399 & 0.597 & 0.332 \tabularnewline\cline{1-5} 
 Informer & 0.344 & 0.393 & 0.608 & 0.344 \tabularnewline\cline{1-5}
 Autoformer & 0.241 & 0.351 & 0.550 & 0.363 \tabularnewline\cline{1-5}
 Pyraformer & 0.478 & 0.471 & 0.606 & 0.338 \tabularnewline\cline{1-5}
 FEDformer & 0.229 & 0.338 & 0.562 & 0.375 \tabularnewline\cline{1-5}
 Crossformer & 0.156 & 0.255 & 0.491 & 0.274 \tabularnewline\cline{1-5}
 \tabulinesep=1.5mm  GSA-Forecaster  & \textbf{0.138} & \textbf{0.233} & \textbf{0.435} & \textbf{0.251} \tabularnewline\cline{1-5}
\hline 
\end{tabu}
\end{table}

\begin{table*}[b]
\caption{\label{tab:gsa-models}GSA-Forecaster variants following the ablation of certain key features in graph sequence attention.}
\centering
\centering
\begin{tabu}{|c|c|c|}
\hline 
Model & Key Features & Implementation Details\tabularnewline
\hline
Model 1 & \makecell{temporal neighborhood, GRU, \\auxiliary information}  & \makecell[l]{Ablate temporal positional encoding from GSA-Forecaster,\\ i.e., set $w_P = 0$ in Equations~\ref{eq:gsa-1} and \ref{eq:gsa-filter}.}\tabularnewline
\hline 
Model 2 & \makecell{temporal neighborhood, GRU} & \makecell[l]{Further ablate auxiliary information from Model 4,\\i.e., set $w_A = 0$ in Equations~\ref{eq:gsa-1} and \ref{eq:gsa-filter}.}\tabularnewline
\hline 
Model 3 & \makecell{temporal neighborhood} & \makecell[l]{Further ablate GRU from Model 3,\\i.e., in Equation~\ref{eq:gsa_decoder_update}, set \begin{small}${\alpha}_{t+k,\,t+k}'^{(h)}$\end{small} $= 0$, \\and substitute \begin{small}${\alpha}_{t+k,\,t+j}'^{(h)}$\end{small} with \begin{small}${\alpha}_{t+k,\,t+j}^{(h)}$\end{small} obtained by Equation~\ref{eq:gsa_attention_score}.}\tabularnewline
\hline 
Model 4 & \makecell{standard attention} & \makecell[l]{Further ablate the temporal neighborhood from Model 2, \\ i.e., set $M=1$ in Equation~\ref{eq:gsa-1} and $M_1 = M_2 = 0$ in Equation~\ref{eq:gsa-filter}.}\tabularnewline
\hline 
\end{tabu}
\end{table*}

\subsection{Ablation Study: Graph Sequence Attention}
\begin{figure}[b]
    \centering
    \begin{minipage}{0.46\textwidth}
    \captionof{table}{\label{tab:gsa_models_performance}Accuracy of GSA-Forecaster variants on Forecasting for the NYC Taxi dataset.}
        \centering
        \begin{tabular}{|c|c|c|}
           \hline 
Model & Overall RMSE & Overall MAPE\tabularnewline
\hline 
GSA-Forecaster & 7.85 & 15.60\% \tabularnewline
\hline
Model 1 & 7.97 & 15.87\% \tabularnewline
\hline 
Model 2 & 8.03 & 15.93\% \tabularnewline
\hline 
Model 3 & 8.12 & 16.11\% \tabularnewline
\hline 
Model 4 & 8.49 & 16.70\% \tabularnewline
\hline 
        \end{tabular}
    \end{minipage}\hfill
    \begin{minipage}{0.42\textwidth}
        \centering
        \includegraphics[width=\textwidth]{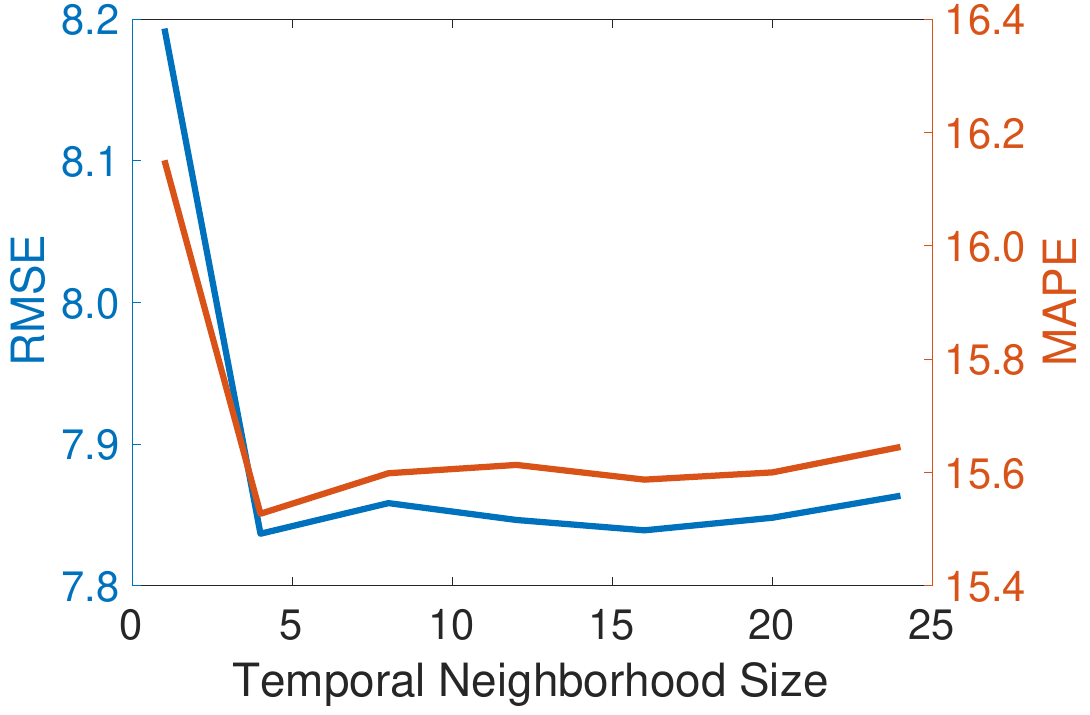} % Replace with your figure file
        \caption{Impact of the temporal neighborhood's size on Forecasting for the NYC Taxi dataset.}
        \label{fig:tm}
    \end{minipage}
\end{figure}

In this study, we sequentially removed features from \mbox{GSA-Forecaster} to understand the impact of each component in graph sequence attention. We started with the full-featured GSA-Forecaster and created successive variants by removing one feature at a time: temporal positional encoding, auxiliary information, GRU, and finally, temporal neighborhood. The specifics of these models are detailed in Table~\ref{tab:gsa-models}, and their overall RMSE and MAPE for forecasting three-hour future taxi demand in New York City are presented in Table~\ref{tab:gsa_models_performance}.

\textbf{Impact of temporal positional encoding:} 
The first variant, Model 1, was created by removing temporal positional encoding from the full GSA-Forecaster. The absence of this feature led to an increase in RMSE and MAPE by 1.57\% and 1.73\%, respectively. This shows the importance of temporal positional encoding in managing data with inherent temporal positional dependencies.

\begin{figure}[t]
    \centering
    \begin{minipage}{0.46\textwidth}
        \centering
        \includegraphics[width=\textwidth]{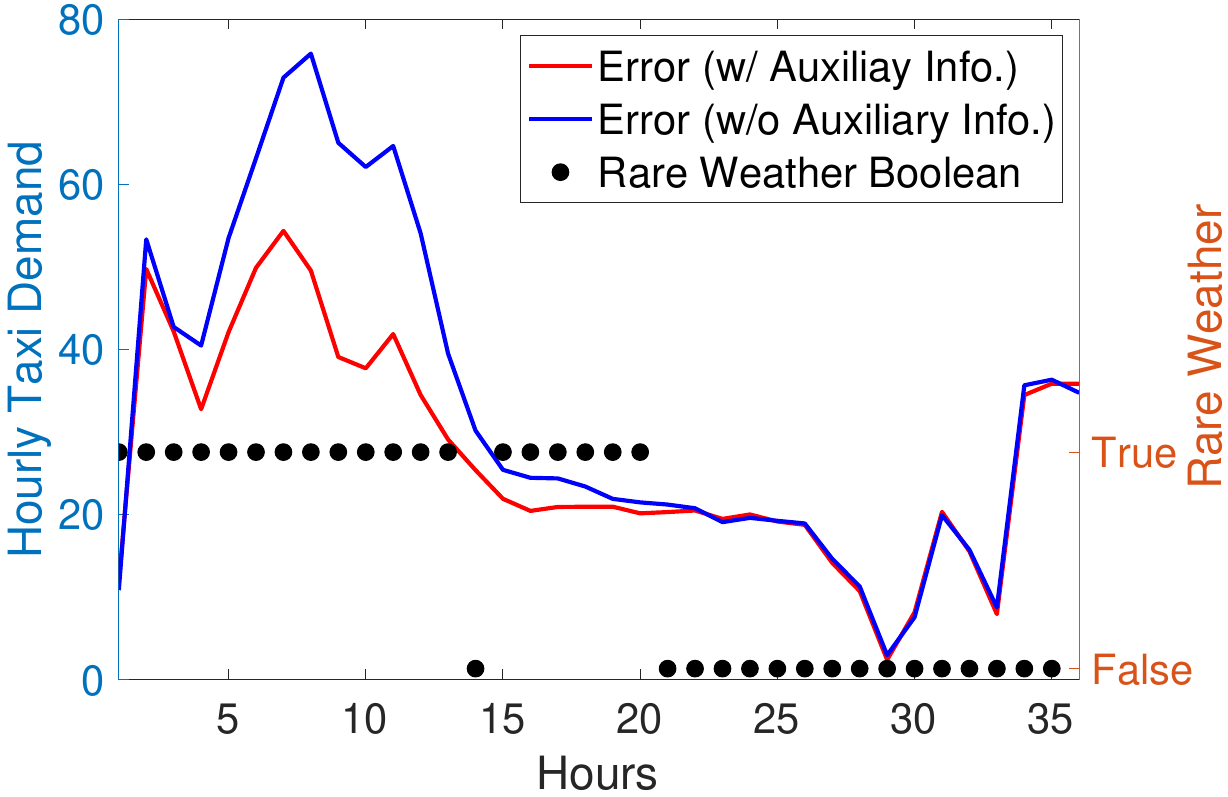} % Replace with your figure file
        \caption{Impact of incorporating auxiliary information on 3-hour-ahead forecasting of taxi pickup demand near Penn Station from January 23, 2016, at 5:00 AM to January 24, 2016, at 5:00 PM.}
        \label{fig:aux_info}
    \end{minipage}\hfill
    \begin{minipage}{0.44\textwidth}
        \centering
        \includegraphics[width=\textwidth]{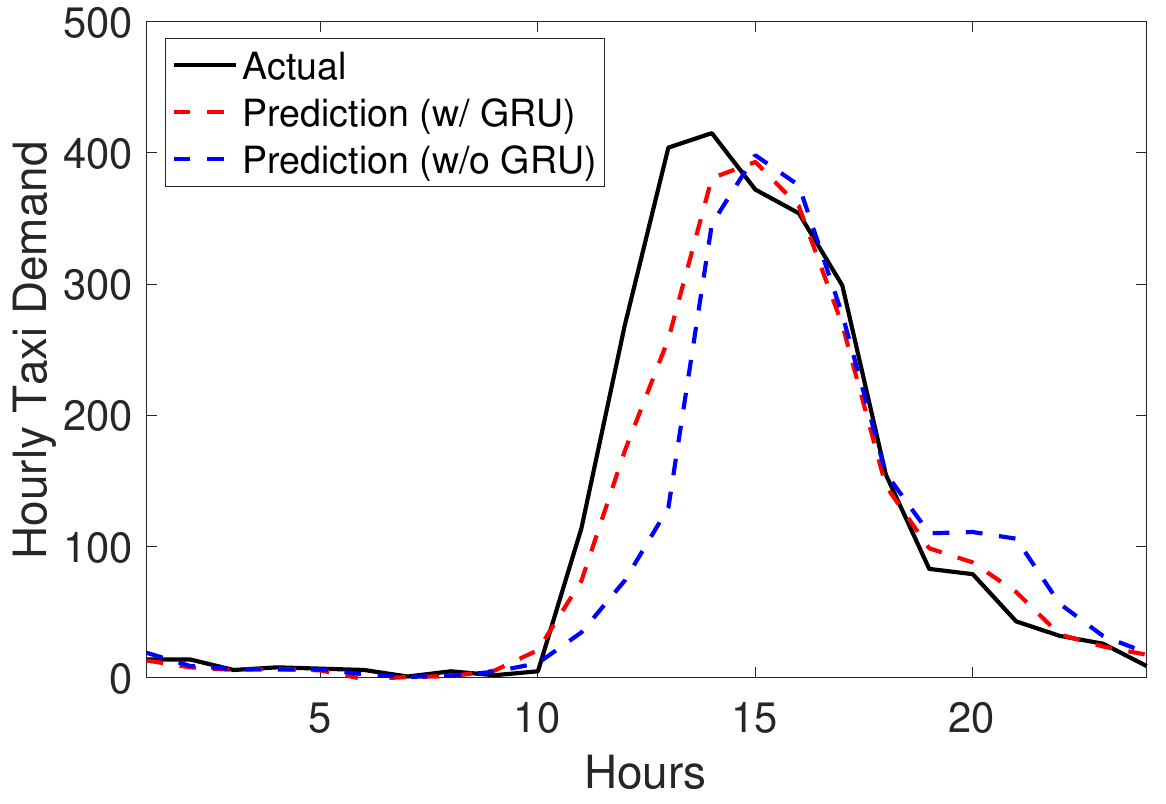} % Replace with your figure file
        \caption{Impact of GRU on 3-hour-ahead forecasting of taxi pickup demand near the Javits Center from November 1, 2015, at 12:00 AM to November 2, 2015, at 12:00 AM.}
        \label{fig:gru_impact}
    \end{minipage}
\end{figure}

\textbf{Impact of auxiliary information:} Next, we removed the auxiliary information from Model 1 to create Model 2. This modification resulted in an increase in RMSE and MAPE by 0.76\% and 0.36\%, respectively, across the entire test set, and a more pronounced increase of 7.0\% and 6.0\% during challenging weather conditions, such as rain or snow. Figure~\ref{fig:aux_info} illustrates the impact of auxiliary information on prediction accuracy. In this example, we perform 3-hour-ahead forecasting of taxi pickup demand near New York Penn Station from January 23, 2016, at 5:00 AM to January 24, 2016, at 5:00 PM. During this period, a severe snowstorm significantly affected taxi pickup demand. The results demonstrate that including auxiliary information, such as weather conditions, substantially improves prediction accuracy during the snowstorm. However, the results also show that incorporating auxiliary information makes little difference once the storm subsides, as expected. This underscores the importance of auxiliary information in capturing the effects of significant external factors, particularly in rare or exceptional scenarios.

\textbf{Impact of GRU:}
Removing the GRU from Model 2 to create Model 3 further degraded the model's performance, resulting in an additional increase in RMSE and MAPE by 1.05\% and 1.11\%, respectively. Figure~\ref{fig:gru_impact} illustrates the prediction results with and without GRU for 3-hour-ahead forecasting of taxi pickup demand near the New York Javits Center from November 1, 2015, at 12:00 AM to November 2, 2015, at 12:00 AM. On November 1, 2015, the Javits Center hosted the 139th International Audio Engineering Society Convention, attracting over 18,500 attendees. This led to significantly higher-than-usual taxi pickup demand, creating non-stationarity in the data, where the historical temporal dependencies leveraged by graph sequence attention were less reliable. The results show that during such non-stationarity, the GRU enables the model to more effectively track recent trends, providing more accurate predictions compared to relying solely on graph sequence attention. This highlights the importance of GRU in capturing recent data patterns, which is crucial in scenarios characterized by strong non-stationarity, where historical data alone is insufficient.

\textbf{Impact of Temporal Neighborhood:}
The removal of the temporal neighborhood feature in Model 4 led to a significant increase in RMSE and MAPE by 4.40\% and 3.53\%, respectively, compared to Model 3. This underscores the crucial role of the temporal neighborhood in graph sequence attention, particularly in its ability to capture temporal dependencies more effectively, offering a substantial improvement over standard attention mechanisms. Figure~\ref{fig:tm} illustrates the GSA-Forecaster's performance across different temporal neighborhood sizes. When the temporal neighborhood size is 1, the model's performance is relatively low. However, as the temporal neighborhood size increases to 4 and beyond, the model's performance improves and stabilizes.

\section{\label{sec:related}Related Work}
Graph-based time-dependent data comprises sets of interdependent time series, exhibiting both within-series (temporal) and between-series (spatial) dependencies. This data is represented using graphs where nodes correspond to individual time series and edges signify spatial dependencies. When each time series is associated with a physical location, the data is often referred to as spatial time series or spatial- and time-dependent data~\cite{Spatial_Time_Series_Def_2003, Spatial_Time_Series_Def_2019, Forecaster}. It is important to distinguish this from temporal graphs~\cite{temporal_graph}, where nodes do not represent time series and only graph edges evolve over time.

In the context of forecasting graph-based time-dependent data, we introduce a \textit{new} attention mechanism (graph sequence attention) and a \textit{new} model (GSA-Forecaster). These offer significantly enhanced accuracy over existing state-of-the-art models. This section reviews previous models and attention mechanisms, highlighting how our work differs from them.

Traditional time series predictors such as ARIMA~\cite{VAR}, VAR~\cite{VAR}, and causal graph processes~\cite{CausalGraphProcess} often rely on stationarity assumptions, which real-world data frequently violates. In contrast, deep learning models~\cite{DMVSTNET_AAAI2018,STResNet_AAAI17,STGCN_IJCAI2018,STMGCN_AAAI2019,Yao_AAAI2019,DCRNN_ICLR2018, jain2016structural,Transformer,cai2020traffic,Forecaster, zhang1, zhang2, zhang3, GraphWaveNet, GMAN, AGCRN} do not make such assumptions and have gained popularity in forecasting graph-based time-dependent data.

Most deep learning models~\cite{DMVSTNET_AAAI2018,STResNet_AAAI17,STGCN_IJCAI2018,STMGCN_AAAI2019,Yao_AAAI2019,DCRNN_ICLR2018, jain2016structural, GraphWaveNet, AGCRN, zhang2} employ RNN or CNN to extract temporal dependency. However, they struggle with long-range temporal dependencies due to the growing number of operations needed to relate data at distant positions\cite{Transformer}. Attention mechanisms, pioneered in NLP and later adopted in various fields, directly relate distant data points in a sequence, effectively capturing long-range dependencies. The Transformer model\cite{Transformer} and its extensions~\cite{OpenAI_Transformer,Transformer_XL,XLNet,Bert,koncel,GPT3,GPT4, ChatGPT} have successfully employed attention mechanisms in tasks such as machine translation and language modeling. These mechanisms have since been applied to other domains, including computer vision~\cite{Image_Transformer, Sparse_Transformer,NonLocal_Net, transformer_object_detection, ViT}, graph learning~\cite{chen2019path,NIPS2019_Korea,GAT}, automated coding~\cite{alphacode, codex}, and time series forecasting~\cite{Forecaster,ucsb_nips19,cai2020traffic,GMAN, triformer, difussion}. Forecaster~\cite{Forecaster}, a model for forecasting graph-based time-dependent data, is built on Transformer's architecture, leveraging its capacity of modeling temporal dependency (though not sufficient for graph-based time-dependent data) while also learning spatial dependencies through sparse linear layers reflecting the data's graph structure.

Most existing models~\cite{Transformer,OpenAI_Transformer,Transformer_XL,XLNet,Bert,koncel,GPT3,GPT4,ChatGPT,ViT,Image_Transformer,Sparse_Transformer,NonLocal_Net,Forecaster,cai2020traffic,chen2019path,NIPS2019_Korea,GAT,GMAN} use standard attention mechanisms, which only compare data encodings at two sequence positions. Our research shows that this approach is inadequate for capturing temporal dependencies in graph-based time-dependent data. Unlike standard attention, our graph sequence attention considers temporal neighborhoods, enhancing the capture of temporal dependencies (see Figure~\ref{fig:GSA_Result} for an example). Our method directly compares data encodings across temporal neighborhoods to capture dependencies, avoiding the potential accuracy loss associated with convolution-based feature aggregation used in Li et al.~\cite{ucsb_nips19}.

For spatial dependencies, prior research~\cite{zhang1, GraphWaveNet, GMAN, DCRNN_ICLR2018, AGCRN, Forecaster, graph_new1, graph_new2, graph_new3, graph_new4, graph_new5, graph_new6} often employs graph neural networks. GSA-Forecaster adopts Forecaster's~\cite{Forecaster} approach of using sparse linear layers to reflect spatial dependencies. Our model uses Gaussian Markov Random Fields to construct a spatial dependency graph when such a graph is not provided. All linear layers, including those within the attention mechanism, are sparse and reflect the graph structure, preserving spatial relationships throughout the model.

Finally, our approach differs from Crossformer~\cite{crossformer}, a recently proposed model leveraging temporal neighborhoods for time series forecasting. For spatial dependencies, we use Gaussian Markov Random Fields to explicitly construct and model spatial relationships through sparse layers, which Crossformer does not address. For temporal dependencies, while both approaches use temporal neighborhoods, Crossformer aggregates embeddings for each neighborhood, potentially leading to accuracy loss. In contrast, our method allows embeddings at different time points to directly interact, avoiding aggregation limitations. Additionally, we incorporate auxiliary information and temporal positional embeddings, further enhancing robustness and accuracy in dynamic scenarios. These distinctions highlight the broader applicability and superior performance of GSA-Forecaster.\vspace{-0.2em}

\section{\label{sec:conclusion}Conclusion}
Forecasting graph-based time-dependent data offers significant advantages in various applications. Accurate forecasting requires models to capture spatial and temporal dependencies effectively and incorporate valuable auxiliary information. In this paper, we introduce GSA-Forecaster, a new deep learning model designed for forecasting graph-based time-dependent data. GSA-Forecaster utilizes a novel graph sequence attention mechanism, which we propose to capture temporal dependency more effectively than the standard attention mechanism prevalent in earlier models. Additionally, \mbox{GSA-Forecaster} integrates the graph structure into its architecture, addressing spatial dependency, and also incorporates auxiliary information to enhance forecasting accuracy.

We applied GSA-Forecaster to a large-scale, real-world graph-based time-dependent dataset focusing on hourly taxi demand in New York City. Our evaluation shows that \mbox{GSA-Forecaster} achieves significant improvements over state-of-the-art predictors including DCRNN, Graph WaveNet, Transformer, and Forecaster, reducing RMSE by 6.7\% and MAPE by 5.8\%. Furthermore, GSA-Forecaster has been successfully applied to forecasting traffic speed, road occupancy rates, and electricity consumption loads, where it consistently outperforms existing state-of-the-art models.

\begin{acks}
This work was partially supported by the Division of Computing and Communication Foundations of the National Science Foundation (Award CCF-2327905). We thank Mr. Kausshik Manojkumar from Iowa State University for his assistance with the experiments.
\end{acks}
\bibliographystyle{ACM-Reference-Format}
\bibliography{reference}
\end{document}